\documentclass{article}

    \PassOptionsToPackage{numbers, compress}{natbib}

    \usepackage[preprint]{neurips_2022}

\usepackage[utf8]{inputenc} 
\usepackage[T1]{fontenc}    
\usepackage{hyperref}       
\usepackage{url}            
\usepackage{booktabs}       
\usepackage{amsfonts}       
\usepackage{nicefrac}       
\usepackage{microtype}      
\usepackage{xcolor}         
\usepackage{amsmath}
\usepackage{amssymb}
\usepackage{graphicx}
\usepackage{caption}
\usepackage{tabularx}
\usepackage{breqn}
\usepackage{flafter}
\usepackage{subcaption}
\usepackage{float}
\usepackage{makecell}
\usepackage{enumitem}
\usepackage{multirow}
\usepackage{array}
\usepackage[flushleft]{threeparttable}

\usepackage{algorithm}
\usepackage{algpseudocode}

\newcolumntype{C}[1]{>{\centering\arraybackslash}p{#1}}
\newcolumntype{P}[1]{>{\centering\arraybackslash}m{#1}}
\newcolumntype{L}[1]{>{\arraybackslash}m{#1}}

\newcommand{\IE}{\mathbb{E}}
\newcommand{\EL}{\mathcal{L}}
\newcommand{\bx}{\mathbf{x}}
\newcommand{\bxt}{\mathbf{x}_{target}}
\newcommand{\by}{\mathbf{y}}
\DeclareMathOperator{\sign}{sign}

\newcommand{\EF}{\mathcal{F}}
\newcommand{\norm}[1]{\left\lVert#1\right\rVert}

\newcommand\crule[3][black]{\textcolor{#1}{\rule{#2}{#3}}}
\DeclareRobustCommand{\eg} {\textit{e}.\textit{g}.}
\DeclareRobustCommand{\ie}{\textit{i}.\textit{e}.}

\DeclareRobustCommand{\NAME}{\textit{AdvBiom}~}
\DeclareRobustCommand{\NAMENOSPACE}{\textit{AdvBiom}}

\algblock{Input}{EndInput}
\algnotext{EndInput}
\algblock{Output}{EndOutput}
\algnotext{EndOutput}
\newcommand{\Desc}[2]{\State \makebox[2em][l]{#1}#2}

\newcommand{\ra}[1]{\renewcommand{\arraystretch}{#1}}

\title{AdvBiom: Adversarial Attacks on Biometric Matchers}
\author{Debayan Deb, Vishesh Mistry, Rahul Parthe\\
TECH5,\\
Troy, MI, USA\\
{\tt\small{\{debayan.deb, vishesh.mistry, rahul.parthe\}@tech5-sa.com}}}

\begin{document}

\maketitle

\begin{abstract}
  With the advent of deep learning models, face recognition systems have achieved impressive recognition rates. The workhorses behind this success are Convolutional Neural Networks (CNNs) and the availability of large training datasets. However, we show that small human-imperceptible changes to face samples can evade most prevailing face recognition systems. Even more alarming is the fact that the same generator can be extended to other traits in the future. In this work, we present how such a generator can be trained and also extended to other biometric modalities, such as fingerprint recognition systems.
\end{abstract}

\section{Introduction}

The last decade has seen a massive influx of deep learning-based technologies that have tackled problems which were once thought to be unsolvable. Much of this progress can be attributed to Convolutional Neural Networks (CNNs)~\cite{cnn, cnn_imagenet} which are now deployed in a plethora of applications ranging from cancer detection to driving autonomous vehicles. Akin to the computer vision domain, the use of CNNs have completely changed the face of biometrics due to the availability of powerful computing devices (GPUs, TPUs) and deep architectures capable of learning rich features~\cite{arcface, sphereface, facenet}. Automated face recognition systems (AFR) have been proven to achieve accuracies as high as 99\% True Accept Rate (TAR) @ 0.1\% False Accept Rate (FAR)~\cite{face_far_tar}, majorly owing to publicly available large-scale face datasets.

Unfortunately, studies have shown that CNN-based networks are vulnerable to adversarial perturbations\footnote{Adversarial perturbations refer to altering an input image instance with small, human imperceptible changes in a manner that can evade CNN models.}~\cite{cnn_adv1, cnn_adv2, cnn_adv3, cnn_adv4, cnn_adv5, cnn_adv6}. It is not surprising that AFR systems too are not impervious to these attacks. Adversarial attacks to an AFR system can be classified into two categories - (i) \emph{impersonation attack} where the hacker tries to perturb his face image to match it to a target victim, and (ii) \emph{obfuscation attack} where the hacker's face image is perturbed to match with a random identity. Both the above attacks involve the hacker adding targeted human-imperceptible perturbations to the face image. These adversarial attacks are different from \emph{face digital manipulation} that include attribute manipulation and synthetic faces, and also from \emph{presentation attacks} which involves the perpetrator wearing a physical artifact such as a mask or replaying a photograph/video of a genuine individual which may be conspicuous in scenarios where human operators are involved.

Let us consider, as example, the largest deployment of fingerprint recognition systems - India's Aadhaar Project~\cite{aadhaar}, which currently has an enrolled gallery size of about 1.35 billion faces from nearly all of its citizens. In September 2022 alone, Aadhaar received ~1.3 billion authentication requests\footnote{\url{https://bit.ly/3BzlpZJ}}. In order to deny a citizen his/her rightful access to government benefits, healthcare, and financial services, an attacker can maliciously perturb enrolled face images such that they do not match to the genuine person during verification. In a typical AFR system, adversarial faces can be replaced with a captured face image in order to prevent the probe face from matching to any of its corresponding enrolled faces. Additionally, the attacker can compromise the entire gallery by inserting adversarial faces in the enrolled gallery, where no probe face will match to the correct identity's gallery.

Adversarial attacks can further be categorized into two types of attacks based on how the attack vector is trained and generated:

\begin{enumerate}
    \item \textit{White-box} attack: Attacks in which the hacker has full knowledge of the recognition system, and iteratively perturbs every pixel by various optimization schemes are termed as white-box attacks~\cite{fgsm, ifgsm, deepfool, pgd, eval_robust_nn, spat_trans_adv, rob_attack_dl, lim_dl_adv, adv_ml}.
    \item \textit{Black-box} attack: With no information about the parameters of the recognition system, black-box attacks are deployed by 
    either transferring attacks learned from an available AFR system~\cite{mifgsm, input_diversity, advfaces, Delv_transf_adv, aifgsm, tifgsm}, or querying the the target system for score~\cite{score_based_1, score_based_2, score_based_3} or decision~\cite{blackbox_decision, boundary_attack}.
    \item \emph{Semi-whitebox} attack: Here, a white-box model is utilized \emph{only during training} and then adversarial examples are synthesized during inference without any knowledge of the deployed AFR model. 
\end{enumerate}




We propose an automated adversarial synthesis method, named \NAMENOSPACE, which generates an adversarial image for a probe image and satisfies all the above requirements. The contributions of the paper are as follows:
\begin{enumerate}
    \item GAN-based \NAME that learns to generate visually realistic adversarial face images that are misclassified by state-of-the-art automated biometric systems.
    \item Adversarial images generated via \NAME are model-agnostic and transferable, and achieve high success rate on 5 state-of-the-art automated face recognition systems. 
    \item Visualizing regions where pixels are perturbed and analyzing the transferability of \NAME.
    \item We show that \NAME achieves significantly higher attack success rate under current defense mechanisms compared to baselines.
    \item With the addition of the proposed Minutiae Displacement and Distortion modules, we show that\NAME can also be extended to successfully evade automated fingerprint recognition systems.
\end{enumerate}


\section{Related Work}

\subsection{Adversarial Attacks}

As discussed earlier, adversarial attacks are broadly classified into \textit{white-box} attacks and \textit{black-box} attacks. A large number of white-box attacks are gradient-based where they analyze the gradients during the back-propagation of an available face recognition system and perform pixel-wise perturbations to the target face image. While approaches such as FGSM~\cite{fgsm} and PGD~\cite{pgd} exploit the high-dimensional space of deep networks to generate adversarial attacks, C\&W~\cite{eval_robust_nn} focuses on minimizing objective functions for optimal adversarial perturbations. However, the basic assumption in white-box attacks that the target recognition system will be available is not plausible. In real-life scenarios, the hacker will not have any information regarding the architecture, training and deployment of the recognition system.

Black-box attacks can be classified into three major categories: transfer-based, score-based, and decision-based attacks. Transfer-based attacks train their adversarial attack generator using readily available recognition systems and then deploy the attacks onto a black-box target system. Dong \textit{et al.}~\cite{mifgsm} proposed the use of momentum for efficient transferability of the adversarial samples. DI$^{2}$-FGSM~\cite{input_diversity} suggested to increase input data diversity for improving transferability. Other approaches in this category include AI-FGSM~\cite{aifgsm} and TI-FGSM~\cite{tifgsm}. Score-based attacks~\cite{score_based_1, score_based_2, score_based_3} query the target system for scores and try to estimate its gradients. Decision-based attacks have the most challenging setting wherein only the decisions from the target system are queried. Some effective methods in this category include Evolutionary attack~\cite{blackbox_decision} and Boundary attack~\cite{boundary_attack}.

\subsection{Adversarial Attacks on Face Recognition}

Although adversarial attacks on face recognition systems have only been recently explored, there has been a significant number of effective approaches for evading AFR systems. Attacks on face recognition systems can be broadly categorized into physical attacks and digital attacks. Physical attacks involve generating adversarial physical artifacts which are 'worn' on a face. Sharif \textit{et al.}~\cite{physical_glass_1, physical_glass_2} proposed generating adversarial eye-glass frames for attacking face recognition systems. In~\cite{advhat}, adversarial printed stickers placed on a hat were generated. However, methods~\cite{physical_glass_1, physical_glass_2, advhat} are implemented in a white-box setting which is unrealistic. Additionally, Nguyen \textit{et al.}~\cite{adv_light} proposed an adversarial light projection attack using an on-premise projector. Yin \textit{et al.}~\cite{adv_makeup} generated and printed eye makeup patches to be stuck around the eyes. More recently, authors in~\cite{rstam} proposed an adversarial mask for impersonation attacks in a black-box setting. However, all the above methods suffer a major drawback of being unrealistic in an operational setting where a human operator is present.

Digital attacks refer to manipulating and perturbing the pixels of a digital face image before being passed through a face recognition system. Early works~\cite{cnn_adv3, eval_robust_nn, cnn_adv4, cnn_adv2, bose} focused on gradient-based attacks for face recognition. However, these methods implement $l_{p}$-norm perturbations to each pixel resulting in decreased attack transferability, and vulnerability to denoising models. Cauli \textit{et al.}~\cite{backdoor_attack} implemented a backdoor attack where the target face recognition system's training samples were manipulated. Apart from the fact that gaining access to the target AFR's training samples is highly improbable, a thorough visual inspection of the samples can easily identify the digital artifacts. Other works employ more stealthy attack approaches against face recognition models. Dong \textit{et al.}~\cite{blackbox_decision} proposed an evolutionary optimization method for generating adversarial faces in decision-based black-box settings. However, they require a minimum of 1,000 queries to the target face recognition system before a realistic adversarial face can be synthesized. \cite{adv_attention} added a conditional variational autoencoder and attention modules to generate adversarial faces in a transfer-based black-box setting. However, they solely focused on impersonation attacks and require at least 5 image samples of the target subject for training and inference. Zhong \textit{et al.}~\cite{adv_dropout} implemented dropout~\cite{dropout} to improve the transferability of the adversarial examples. \cite{adv_makeup} perturbed the eye region of a face to produce adversarial eyeshadow artifacts. However, the artifacts are visibly conspicuous under close inspection. Deb \textit{et al.}~\cite{advfaces} used a GAN to generate minimal perturbations in salient facial regions. More recently, \cite{adv_attribute_1} and~\cite{adv_attribute_2} have focused on manipulating facial attributes for targeted adversarial attacks.


\section{Adversarial Faces}

\subsection{Preliminaries}
The goal of any attacker is to evade Automated Face Recognition (AFR) systems under either of the two settings:
\begin{itemize}
\item \textbf{Obfuscation} Manipulate input face images in a manner such that they cannot be identified as the hacker, or
\item \textbf{Impersonation} Edit input face images such that they are identified as a target/desired individual (victim).
\end{itemize}
While the manipulated face image evades the AFR system, a key requirement in a successful attack is such that the input face image should appear as a legitimate face photo of the attacker. In 
other words, the attacker desires an automated method of adding small and human-imperceptible changes to an input face image such that it can evade AFR systems while appear benign to human observers. These changes are denoted as \emph{adversarial perturbations} and the manipulated image is hereby referred to as \emph{adversarial images}\footnote{We interchangeably use the terms adversarial images and adversarial faces in this paper.}. In addition, the automated method of synthesizing adversarial perturbations is named as \emph{adversarial generator}.

Formally, given an input face image, $\bx$, an adversarial generator has two requirements under the \emph{obfuscation} scenario:
\begin{itemize}
\item synthesize an adversarial face image, $\bx_{adv} = \bx + \delta$, such that AFR systems fail to match $\bx_{adv}$ and $\bx$, and
\item limit the magnitude of perturbation $||\delta||_p$ such that $\bx_{adv}$ appears very similar to $\bx$ to humans.
\end{itemize}
When the attack aims to impersonate a target individual, we need an image of the victim $\bxt$ where the identity of $\bx$ and $\bxt$ are different. Therefore, constraints under the \emph{impersonation} setting are as follows:
\begin{itemize}
\item synthesize an adversarial face image, $\bx_{adv} = \bx + \delta$, such that AFR systems erroneously match $\bx_{adv}$ and $\bxt$, and
\item limit the magnitude of perturbation $||\delta||_p$ such that $\bx_{adv}$ appears very similar to $\bx$ to humans.
\end{itemize}

Obfuscation attempts (faces are perturbed such that they cannot be identified as the attacker) are generally more effective~\cite{advfaces}, computationally efficient to synthesize~\cite{fgsm, pgd}, and widely adopted~\cite{fawkes} compared to impersonation attacks (perturbed faces can automatically match to a target subject). Therefore, this paper focuses on crafting obfuscation attacks, however, we will still show examples on synthesizing impersonation attacks.

\subsection{Gradient-based Attacks}
In white-box attacks, the attacker is assumed to have the knowledge and access to the AFR system's model and parameters. Naturally, we then expect a much better attack success rate under white-box settings since the attacker can carefully craft adversarial perturbations that necessarily evade the target AFR system. However, these white-box manipulations of face recognition models are impractical in real-world scenarios. For instance, assuming access to an airport's already deployed AFR system may be extremely difficult.

Nevertheless, it is advantageous to understand prevailing white-box methods. That is, \emph{if given access to a CNN-based AFR system, how could one utilize all of its model parameters to launch a successful adversarial attack?}

A common approach is to utilize gradients of the whitebox AFR models. Namely the attackers modify the image in the direction of the gradient of the loss function with respect to the input image.
There are two prevailing approaches to perform such gradient-based attacks: 
\begin{itemize}
\item \emph{one-shot} attacks, in which the attacker takes a single step in the direction of the gradient, and 
\item \emph{iterative} attacks where instead of a single step, several steps are taken until we obtain a successful adversarial pattern. 
\end{itemize}

\subsubsection{Fast Gradient Sign Method (FGSM)}
This method computes an adversarial image by adding a pixel-wide perturbation of magnitude in the direction of the gradient~\cite{fgsm}. Under FGSM attack, we take a single step towards the direction of the gradient, and therefore, FGSM is very efficient in terms of computation time. Formally, given an input image $\bx$, we obtain an adversarial image $\bx_{adv}$:
\begin{align*}
\bx_{adv} = \bx + \epsilon \cdot \sign \left(\triangledown_{\bx}\text{J}\left(\bx, \mathbf{y} \right) \right)
\end{align*}
\noindent where, $\text{J}$ is the loss function used to train the AFR system (typically, softmax cross entropy loss), and $\mathbf{y}$ is the ground truth class label of $\bx$ (typically, the subject ID of the identity in $\bx$).

FGSM was first proposed for the object classification domain and therefore, utilizes softmax probabilities for crafting adversarial perturbations. Therefore, the number of object classes are assumed to be known during training and testing. However, face recognition systems do not utilize the softmax layer for classification (as the number of identities are not fixed during deployment) instead features from the last fully connected layer are used for comparing face images. 

We first modify FGSM appropriately in order to evade AFR systems rather than object classifiers. Instead of considering the softmax cross-entropy loss as $\text{J}$, we craft a new loss function that models real-world scenario\footnote{For brevity, we denote $\IE_\bx \equiv \IE_{\bx \in \mathcal{P}_{data}}$.}:
\begin{align*}
\EL_{featureMatch} = 1 -\IE_{\bx}\left[\frac{\EF(\bx) \cdot \EF(\bx_{adv})}{\left|\left|\EF(\bx)\right|\right|\left|\left|\EF(\bx_{adv})\right|\right|}\right].
\end{align*}
\noindent where, $\EF$ is the matcher and $\EF(\bx)$ is the feature representation of an input image $\bx$. The above feature matching loss function computes the cosine distance between a pair of images and ensures that the features between adversarial image $\bx_{adv}$ and input image $\bx$ are as close as possible. Therefore, the gradient of the above loss ensures the features do not match and hence, can be considered as an obfuscation adversarial attack.

\begin{figure}[h!]
    \centering
    \captionsetup{font=footnotesize}
    \subfloat[Real Input Image]{\includegraphics[width=0.18\textwidth]{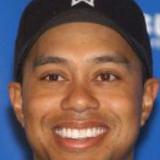}}~~~~~
    \subfloat[Perturbation]{\includegraphics[width=0.18\textwidth]{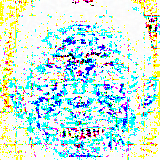}}~~~~~
    \subfloat[FGSM~\cite{fgsm}]{\includegraphics[width=0.18\textwidth]{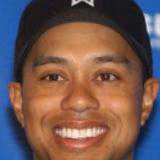}}
    \caption{Adversarial face synthesized via FGSM \cite{fgsm}. A state-of-the-art face matcher, ArcFace \cite{arcface}, fails to match the adversarial and input image. Cosine similarity score ($\in[-1,1]$) between the two images is $\textbf{0.27}$, while a score above $\textbf{0.36}$ (threshold @ $0.1\%$ False Accept Rate) indicates that two faces are of the same subject.}
    \label{fig:fgsm_set}
\end{figure}

In Fig. \ref{fig:fgsm_set}, we show the results of launching our modified FGSM attack on a state-of-the-art AFR system, namely ArcFace \cite{arcface}. We see that with a single step and with minimal perturbations, the real and adversarial images of Tiger Woods does not match via ArcFace while humans can easily identity both images as pertaining to the same subject.

\subsubsection{Projected Gradient Descent (PGD)}
\begin{figure}[h!]
    \centering
    \captionsetup{font=footnotesize}
    \subfloat[Real Input Image]{\includegraphics[width=0.18\textwidth]{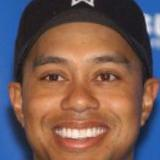}}~~~~~
    \subfloat[Perturbation]{\includegraphics[width=0.18\textwidth]{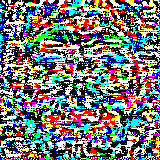}}~~~~~
    \subfloat[PGD~\cite{pgd}]{\includegraphics[width=0.18\textwidth]{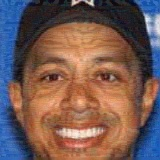}}
    \caption{Adversarial face synthesized via PGD \cite{pgd}. A state-of-the-art face matcher, ArcFace \cite{arcface}, fails to match the adversarial and input image. Cosine similarity score ($\in[-1,1]$) between the two images is $\textbf{0.12}$, while a score above $\textbf{0.36}$ (threshold @ $0.1\%$ False Accept Rate) indicates that two faces are of the same subject.}
    \label{fig:pgd_set}
\end{figure}

An extreme case of white-box attacks is the PGD attack \cite{pgd} where we assume that the attacker also has unlimited number of attempts to try and evade the deployed AFR system. Unlike FGSM, PGD is an iterative attack. PGD attempts to find the perturbation $\delta$ that maximises the loss of a model on a particular input while keeping the size of the perturbation smaller than a specified amount referred to as $\epsilon$. We keep iterating until such a $\delta$ is obtained. Similar to FGSM, we modify the loss function of PGD to fit the requirements of AFR system by again considering $\EL_{featureMatch}$ as the loss. Fig. \ref{fig:pgd_set} shows the results of PGD attack on ArcFace matcher. Note that due to multiple iterations, PGD attack on AFR systems is more powerful (lower cosine similarity) but also more visible to humans as compared to the single-step FGSM attack.

\subsection{Geometric Perturbations (GFLM)} Prior efforts in crafting adversarial faces have also tried non-linear deformations as a natural method for evading AFR systems \cite{gflm}. Non-linear deformations are applied by performing geometric warping to the input face images.

\begin{figure}[h!]
    \centering
    \captionsetup{font=footnotesize}
    \subfloat[Real Input Image]{\includegraphics[width=0.18\textwidth]{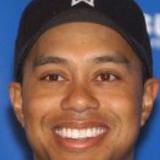}}~~~~~
    \subfloat[Perturbation]{\includegraphics[width=0.18\textwidth]{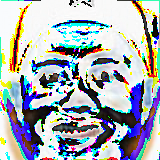}}~~~~~
    \subfloat[GFLM]{\includegraphics[width=0.18\textwidth]{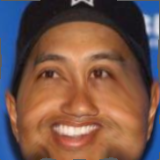}}
    \caption{Adversarial face synthesized via GFLM \cite{gflm}. A state-of-the-art face matcher, ArcFace \cite{arcface}, fails to match the adversarial and input image. Cosine similarity score ($\in[-1,1]$) between the two images is $\textbf{0.33}$, while a score above $\textbf{0.36}$ (threshold @ $0.1\%$ False Accept Rate) indicates that two faces are of the same subject.}
    \label{fig:gflm_set}
\end{figure}

Unlike traditional adversarial perturbations that basically add an adversarial perturbation $\delta$, authors in \cite{gflm} propose a fast method of generating adversarial faces by altering the landmark locations of the input images. The resulting adversarial faces completely lie on the manifold of natural images, which makes it extremely difficult to detect any adversarial perturbations. Results of geometrically warped adversarial faces are presented in \ref{fig:gflm_set}.

\subsection{Attribute-based Perturbations}
Unlike geometric-warping and gradient-based attacks that may perturb every pixel in the image, a few studies propose manipulating only salient regions in faces, {\it e.g.}, eyes, nose, and mouth.

\begin{figure}[h!]
    \centering
    \captionsetup{font=footnotesize}
    
    \subfloat[Real Input Image]{\includegraphics[width=0.15\textwidth]{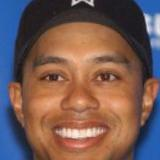}}\hfill
    \subfloat[Blond]{\includegraphics[width=0.15\textwidth]{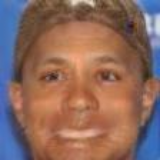}}\hfill
    \subfloat[Bangs]{\includegraphics[width=0.15\textwidth]{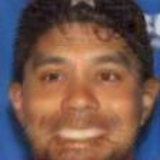}}\hfill
    \subfloat[Mouth Open]{\includegraphics[width=0.15\textwidth]{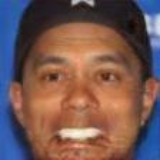}}\hfill
    \subfloat[Eyeglasses]{\includegraphics[width=0.15\textwidth]{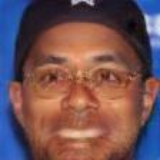}}\hfill
    \subfloat[Makeup]{\includegraphics[width=0.15\textwidth]{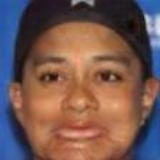}}\\
    \captionsetup[subfigure]{labelformat=empty}
    \subfloat[]{\includegraphics[width=0.15\textwidth]{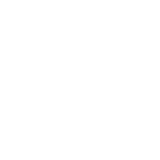}}\hfill
    \subfloat[]{\includegraphics[width=0.15\textwidth]{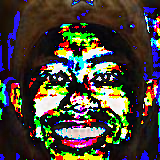}}\hfill
    \subfloat[]{\includegraphics[width=0.15\textwidth]{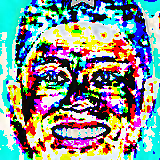}}\hfill
    \subfloat[]{\includegraphics[width=0.15\textwidth]{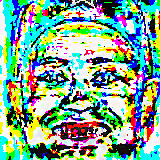}}\hfill
    \subfloat[]{\includegraphics[width=0.15\textwidth]{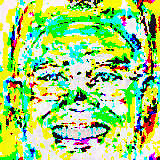}}\hfill
    \subfloat[]{\includegraphics[width=0.15\textwidth]{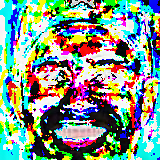}}
    \caption{Adversarial face synthesized via manipulating semantic attributes \cite{adv_attribute_2}. All adversarial images (b-f) fail to match with the real image (a) via ArcFace~\cite{arcface}.}
    \label{fig:semantic_set}
\end{figure}

By restricting perturbations to only semantic regions of the face, SemanticAdv~\cite{adv_attribute_2} generates adversarial examples in a more controllable fashion by editing a single semantic aspect through attribute-conditioned image editing. Fig.~\ref{fig:semantic_set} shows results from adversarial manipulating semantic attributes. We can see while the attacks are indeed successful, it comes at the cost of altering the perceived identity as well as leads to degraded image quality.

\section{\NAMENOSPACE: Learning to Synthesize Adversarial Attacks}

We find that majority of prior efforts on crafting adversarial attacks either degrade the visual quality where an observant human can still visually pick out the adversarial patterns. We also identify the following challenges with prior efforts:
\begin{itemize}
\item \textbf{Gradient-based} attacks rely on white-box settings where the entire deployed CNN-based AFR system is available to the attacker to compute its gradients.
\item \textbf{Geometrically-warping} faces generally do not guarantee adversarial success and greatly distort the face image.
\item \textbf{Semantic attribute} manipulation can also degrade visual quality and may lead to greater conspicuous changes.
\end{itemize}

Instead, we propose to train a network to ``learn" the salient regions of the face that can be perturbed to evade AFR systems in a semi-whitebox setting. These leads to the following advantages over prior efforts:
\begin{itemize}
\item \textbf{Perceptual Realism} Given a large enough training dataset, a network can gradually learn to synthesize adversarial face images that are perceptually realistic such that a human observer can identify the image as a legitimate face image.
\item \textbf{Higher Attack Success}  The faces can be learned to be perturbed in a manner such that they cannot be identified as the hacker (obfuscation at-
tack) or automatically matched to a target subject (impersonation attack) by an AFR system.
\item \textbf{Controllable} The amount of perturbation can also be controllable by
the attacker so that they can examine the success of the learning model as a function of amount of perturbation.
\item \textbf{Transferability} Due to the semi-whitebox setting: once the network learns to generate the perturbed instances based on a single face recognition system, attacks can be transferred to any black-box AFR systems. 
\end{itemize}

We propose an automated adversarial biometric synthesis method, named \NAMENOSPACE, which generates an adversarial image for a probe face image and satisfies all the above requirements.

\subsection{Methodology}

\begin{figure}[h!]
    \centering
    \captionsetup{font=footnotesize}
    \includegraphics[width=\linewidth]{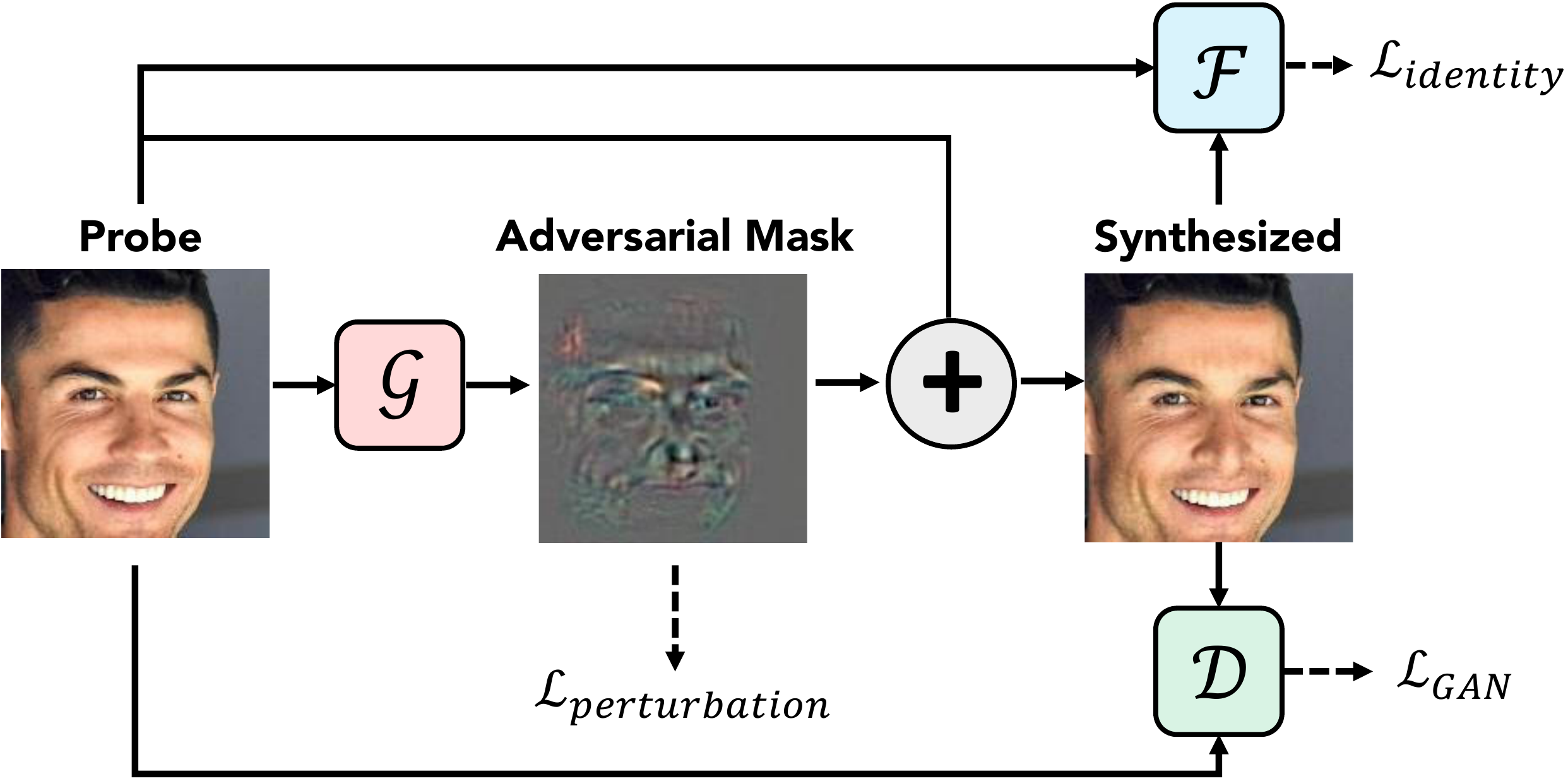}
    \caption{Given a probe face image, \NAME automatically generates an adversarial mask that is then added to the probe to obtain an adversarial face image.}
    \label{fig:arch}
\end{figure}

Our goal is to synthesize a face image that visually appears to pertain to the target face, yet automatic face recognition systems either incorrectly matches the synthesized image to another person or does not match to target's gallery images. \NAME comprises of a generator $\mathcal{G}$, a discriminator $\mathcal{D}$, and face matcher (see Figure~\ref{fig:arch}).

\paragraph{Generator} The proposed generator takes an input face image, $\bx \in \mathcal{X}$, and outputs an image, $\mathcal{G}(\bx)$. The generator is conditioned on the input image $\bx$; for different input faces, we will get different synthesized images.

Since our goal is to obtain an adversarial image that is metrically similar to the probe in the image space, $\bx$, it is not desirable to perturb all the pixels in the probe image. For this reason, we treat the output from the generator as an additive mask and the adversarial face is defined as $\bx + \mathcal{G}(\bx)$. If the magnitude of the pixels in $\mathcal{G}(\bx)$ is minimal, then the adversarial image comprises mostly of the probe $\bx$. Here, we denote $\mathcal{G}(\bx)$ as an ``adversarial mask". In order to bound the magnitude of the adversarial mask, we introduce a \emph{perturbation loss} during training by minimizing the $L_2$ norm\footnote{For brevity, we denote $\IE_\bx \equiv \IE_{\bx \in \mathcal{X}}$.}:
\begin{align}
    \EL_{perturbation} &= \IE_{\bx}\left[\max\left(\epsilon,{\norm{\mathcal{G}(\bx)}_2}\right)\right]
\end{align}
where $\epsilon\in [0,\infty)$ is a hyperparameter that controls the minimum amount of perturbation allowed.

In order to achieve our goal of impersonating a target subject's face or obfuscating one's own identity, we need a face matcher, $\mathcal{F}$, to supervise the training of \NAMENOSPACE. For obfuscation attack, at each training iteration, \NAME tries to minimize the cosine similarity between face embeddings of the input probe $\bx$ and the generated image $\bx + \mathcal{G}(\bx)$ via an \emph{identity} loss function:
\begin{align}
    \EL_{identity} &= \IE_{\bx}{\left[\mathcal{F}(\bx, \bx+\mathcal{G}(\bx))\right]}
\end{align}
For an impersonation attack, \NAME maximizes the cosine similarity between the face embeddings of a randomly chosen target's probe, $\by$, and the generated adversarial face $\bx+\mathcal{G}(\bx)$ via: 
\begin{align}
    \EL_{identity} &= \IE_{\bx}{\left[1- \mathcal{F}(\by, \bx+\mathcal{G}(\bx))\right]}
\end{align}

The perturbation and identity loss functions enforce the network to learn the salient facial regions that can be perturbed minimally in order to evade automatic face recognition systems.

\paragraph{Discriminator}
Akin to previous works on GANs~\cite{gan, image_synthesis_1}, we introduce a discriminator in order to encourage perceptual realism of the generated images. We use a fully-convolution network as a patch-based discriminator~\cite{image_synthesis_1}. Here, the discriminator, $\mathcal{D}$, aims to distinguish between a probe, $\bx$, and a generated adversarial face image $\bx+\mathcal{G}(\bx)$ via a GAN loss:
\begin{align}
    \begin{split}
    \EL_{GAN} = &\quad\IE_{\bx}\left[{\log\mathcal{D}(\bx)}\right] +\\
    &\quad\IE_{\bx}{\left[\log(1-\mathcal{D}(\bx+\mathcal{G}(\bx)))\right]}
    \end{split}
\end{align}

Finally, \NAME is trained in an end-to-end fashion with the following objectives:
\begin{align}
\min_{\mathcal{D}}\EL_{\mathcal{D}} &= -\EL_{GAN}\\
\min_{\mathcal{G}} \EL_{\mathcal{G}} &= \EL_{GAN} + \lambda_{i} \EL_{identity} + \lambda_{p}\EL_{perturbation}
\end{align}
where $\lambda_i$ and $\lambda_p$ are hyper-parameters controlling the relative importance of identity and perturbation losses, respectively. Note that $\EL_{GAN}$ and $\EL_{perturbation}$ encourage the generated images to be visually similar to the original face images, while $\EL_{identity}$ optimizes for a high attack success rate.  
After training, the generator $\mathcal{G}$ can generate an adversarial face image for any input image and can be tested on any black-box face recognition system.

The overall algorithm describing the training procedure of \NAME can be found in Algorithm~\ref{apdx:alg}.

\begin{algorithm}[!ht]
    \caption{Training \NAMENOSPACE. All experiments in this work use $\alpha = 0.0001$, $\beta_1 = 0.5$, $\beta_2 = 0.9$, $\lambda_i = 10.0$, $\lambda_p = 1.0$, $m = 32$. \\We set $\epsilon = 3.0$ (obfuscation), $\epsilon = 8.0$ (impersonation).}\label{alg:advfaces}
  \begin{algorithmic}[1]
     \Input
     \Desc{$X$}{Training Dataset}
      \Desc{$\mathcal{F}$}{Cosine similarity between an image pair obtained by biometric matcher}
      \Desc{$\mathcal{G}$}{Generator with weights $\mathcal{G}_\theta$}
      \Desc{$\mathcal{D}$}{Discriminator with weights $\mathcal{D}_\theta$}
      \Desc{$m$}{Batch size}
      \Desc{$\alpha$}{Learning rate}
  \EndInput
    \For{number of training iterations}
        \State \text{Sample a batch of probes $\{x^{(i)}\}_{i=1}^{m} \sim \mathcal{X}$}
        \If{impersonation attack}
            \State Sample a batch of target images ${y^{(i)}} \sim \mathcal{X}$
            \State $\delta^{(i)} = \mathcal{G}((x^{(i)}, y^{(i)})$
        \ElsIf{obfuscation attack}
            \State $\delta^{(i)} = \mathcal{G}(x^{(i)})$
        \EndIf
        \State $x_{adv}^{(i)} = x^{(i)} + \delta^{(i)}$
        \State $\EL_{perturbation} = \frac{1}{m}\left[ \sum_{i=1}^{m}\max\left(\epsilon,||\delta^{(i)}||_2\right)\right]$
        \If{impersonation attack}
            \State $\EL_{identity} = \frac{1}{m}\left[\sum_{i=1}^{m}\mathcal{F}\left(x^{(i)}, x_{adv}^{(i)}\right)\right]$
        \ElsIf{obfuscation attack}
            \State $\EL_{identity} = \frac{1}{m}\left[\sum_{i=1}^{m}\left(1 - \mathcal{F}\left(y^{(i)}, x_{adv}^{(i)}\right)\right)\right]$
        \EndIf
        \State $\EL^{\mathcal{G}}_{GAN} = \frac{1}{m}\left[\sum_{i=1}^{m}log\left(1-\mathcal{D}(x_{adv}^{(i)}) \right)\right]$
        \State $\EL^{\mathcal{D}} = \frac{1}{m}\sum_{i=1}^{m}\left[log\left(\mathcal{D}(x^{(i)}) \right) +  log\left(1-\mathcal{D}(x_{adv}^{(i)}) \right)\right]$
        \State $\EL^{\mathcal{G}} = \EL^{\mathcal{G}}_{GAN} + \lambda_i \EL_{identity} + \lambda_p \EL_{perturbation}$
        \State $\mathcal{G}_{\theta} = \texttt{Adam}(\triangledown_{\mathcal{G}}\EL^{\mathcal{G}}, \mathcal{G}_{\theta},\alpha,\beta_{1},\beta_{2})$
        \State $\mathcal{D}_\theta = \texttt{Adam}(\triangledown_{\mathcal{D}}\EL^{\mathcal{D}}, \mathcal{D}_{\theta},\alpha,\beta_{1},\beta_{2})$
      \EndFor
    \end{algorithmic}
    \label{apdx:alg}
\end{algorithm}

\subsection{Experimental Results}

\paragraph{Evaluation Metrics} 
We quantify the effectiveness of the adversarial attacks generated by \NAME and other state-of-the-art baselines via (i) \emph{attack success rate} and (ii) \emph{structural similarity (SSIM)}.

The attack success rate for \emph{obfuscation attack} is computed as,
\begin{align}
   \text{Attack Success Rate} &= \frac{\text{(No. of Comparisons $< \tau$)}}{\text{Total No. of Comparisons}}
   \label{eq:attack1}
\end{align}
where each comparison consists of a subject's adversarial probe and an enrollment image. Here, $\tau$ is a pre-determined threshold computed at, say, 0.1\% FAR\footnote{For each face matcher, we pre-compute the threshold at $0.1\%$ FAR on all possible image pairs in LFW. For~\eg, threshold @ $0.1\%$ FAR for ArcFace is $0.28$.}. Attack success rate for \emph{impersonation attack} is defined as,
\begin{align}
   \text{Attack Success Rate} &= \frac{\text{(No. of Comparisons $\geq \tau$)}}{\text{Total No. of Comparisons}}
   \label{eq:attack2}
\end{align}
Here, a comparison comprises of an adversarial image synthesized with a target's probe and matched to the target's enrolled image. We evaluate the success rate for the impersonation setting via 10-fold cross-validation where each fold consists of a randomly chosen target.

Similar to prior studies~\cite{adv_attention}, in order to measure the similarity between the adversarial example and the input face, we compute the structural similarity index (SSIM) between the images. SSIM is a normalized metric between $-1$ (completely different image pairs) to $1$ (identical image pairs).

\paragraph{Datasets} We train \NAME on CASIA-WebFace~\cite{casia} and then test on LFW~\cite{lfw}\footnote{Training on CASIA-WebFace and evaluating on LFW is a common approach in face recognition literature~\cite{arcface, sphereface}}.
\begin{itemize}[topsep=0.1em, itemsep=-0.1em]
    \item \textbf{CASIA-WebFace}~\cite{casia} is comprised of 494,414 face images belonging to 10,575 different subjects. We removed 84 subjects that are also present in LFW and the testing images in this paper.
    \item \textbf{LFW}~\cite{lfw} contains 13,233 web-collected images of 5,749 different subjects. In order to compute the attack success rate, we only consider subjects with at least two face images. After this filtering, 9,614 face images of 1,680 subjects are available for evaluation.
\end{itemize}
\emph{All the testing images in this paper have no identity overlap with the training set, CASIA-WebFace~\cite{casia}.}

\paragraph{Data Preprocessing} All face images are passed through MTCNN face detector~\cite{mtcnn} to detect five landmarks (two eyes, nose, and two mouth corners). Via similarity transformation, the face images are aligned. After transformation, the images are resized to $160\times 160$. Prior to training and testing, each pixel in the RGB image is normalized by subtracting 127.5 and dividing by 128.

\paragraph{Experimental Settings}
We use ADAM optimizers in Tensorflow with $\beta_{1} = 0.5$ and $\beta_{2}=0.9$ for the entire network. Each mini-batch consists of $32$ face images. We train \NAME for 200,000 steps with a fixed learning rate of $0.0001$. Since our goal is to generate adversarial faces with high success rate, the identity loss is of utmost importance. We empirically set $\lambda_i = 10.0$ and $\lambda_p = 1.0$. We train two separate models and set $\epsilon = 3.0$ and $\epsilon = 8.0$ for obfuscation and impersonation attacks, respectively. 

\begin{figure*}[!h]
\captionsetup{font=footnotesize}
\footnotesize
\centering\begin{tabular}{c@{ }c@{ }c@{ }c@{ }c@{ }c@{}}
\textbf{Gallery} & \textbf{Probe} & \textbf{Proposed \NAME} & \textbf{GFLM~\cite{gflm}} &\textbf{PGD~\cite{pgd}} & \textbf{FGSM~\cite{fgsm}} \\
\includegraphics[width=.15\textwidth]{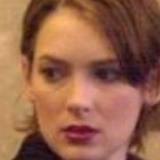}&
\includegraphics[width=.15\textwidth]{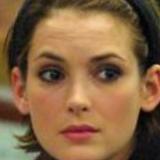}&~~
\includegraphics[width=.15\textwidth]{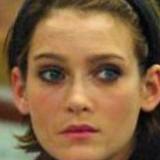}&
\includegraphics[width=.15\textwidth]{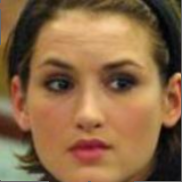}&\includegraphics[width=.15\textwidth]{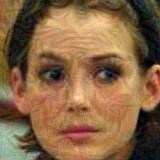} &
\includegraphics[width=.15\textwidth]{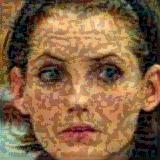}\\
 & 0.68 & 0.14 & 0.26 & 0.27 & 0.04\\
\includegraphics[width=.15\textwidth]{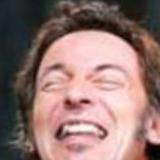}&
\includegraphics[width=.15\textwidth]{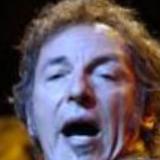}&~~
\includegraphics[width=.15\textwidth]{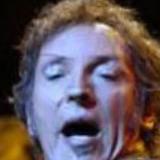}&
\includegraphics[width=.15\textwidth]{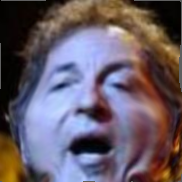}& \includegraphics[width=.15\textwidth]{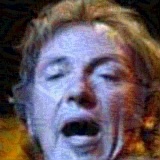} &
\includegraphics[width=.15\textwidth]{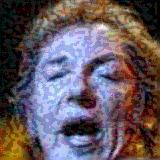}\\
 & 0.38 & 0.08 & 0.12 & 0.21 & 0.02\\
 \end{tabular}\\[0.5em]
(a) Obfuscation Attack\\[1.5em]
\centering\begin{tabular}{c@{ }c@{ }c@{ }c@{ }c@{ }c@{}}
\textbf{Target's Gallery} & \textbf{Target's Probe} & \textbf{Probe} & \textbf{Proposed \NAME} &\textbf{A$^3$GN~\cite{adv_attention}} & \textbf{FGSM~\cite{fgsm}} \\
\includegraphics[width=.15\textwidth]{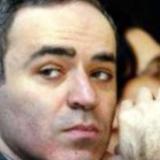}&
\includegraphics[width=.15\textwidth]{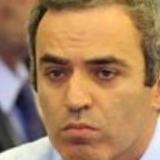}&
\includegraphics[width=.15\textwidth]{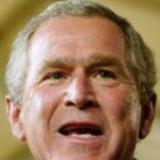} &~~
\includegraphics[width=.15\textwidth]{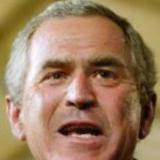}&
\includegraphics[width=.15\textwidth]{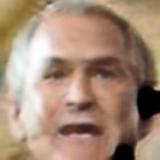}&
\includegraphics[width=.15\textwidth]{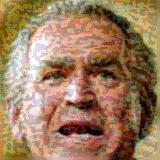}\\
 & 0.78 & 0.10 & 0.30 & 0.29 & 0.36\\
\includegraphics[width=.15\textwidth]{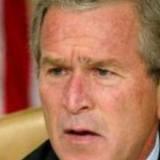}&
\includegraphics[width=.15\textwidth]{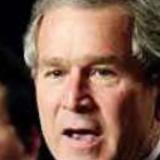}&
\includegraphics[width=.15\textwidth]{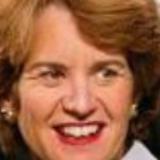}&~~
\includegraphics[width=.15\textwidth]{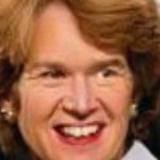}& \includegraphics[width=.15\textwidth]{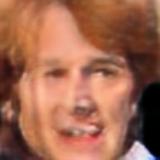}&
\includegraphics[width=.15\textwidth]{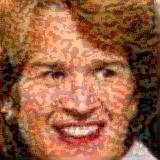}\\
 & 0.80 & 0.15 & 0.34 & 0.33 & 0.42\\
 \end{tabular}\\[0.5em]
 (b) Impersonation Attack%
\caption{Adversarial face synthesis results on LFW dataset in (a) obfuscation and (b) impersonation attack settings (cosine similarity scores obtained from ArcFace~\cite{arcface} with threshold @ $0.1\%$ FAR$=0.28$). The proposed method synthesizes adversarial faces that are seemingly inconspicuous and maintain high perceptual quality.}
\label{fig:comparison}
\end{figure*}

\paragraph{Architecture} Let \texttt{c7s1-k} be a $7\times7$ convolutional layer with $k$ filters and stride $1$. \texttt{dk} denotes a $4\times 4$ convolutional layer with $k$ filters and stride $2$. \texttt{Rk} denotes a residual block that contains two $3\times 3$ convolutional layers. \texttt{uk} denotes a $2\times$
upsampling layer followed by a $5\times 5$ convolutional layer with $k$ filters and stride $1$. We apply Instance Normalization and Batch Normalization to the generator and discriminator, respectively. We use Leaky ReLU with slope 0.2 in the discriminator and ReLU activation in the generator. The architectures of the two modules are as follows:
\begin{itemize}
    \itemsep0em
    \item Generator: \\ 
    \texttt{c7s1-64,d128,d256,R256,R256,R256, u128, u64, c7s1-3}
    \item Discriminator: \\ \texttt{d32,d64,d128,d256,d512}
\end{itemize}
A $1\times 1$ convolutional layer with $3$ filters and stride $1$ is attached to the last convolutional layer of the discriminator for the patch-based GAN loss $\EL_{GAN}$.

\begin{table*}[!t]
\centering
\ra{1.0}
\footnotesize
\resizebox{\textwidth}{!}{
\captionsetup{font=footnotesize}
\begin{threeparttable}
\begin{tabular}{@{}lcccccccc@{}}
\toprule \textbf{Obfuscation Attack} & \multicolumn{1}{c}{Proposed \NAME} & \phantom{abc}& \multicolumn{1}{c}{GFLM~\cite{gflm}} & 
\phantom{abc}& \multicolumn{1}{c}{PGD~\cite{pgd}} & 
\phantom{abc} &\multicolumn{1}{c}{FGSM~\cite{fgsm}}\\
\midrule
Attack Success Rate (\%) @ 0.1\% FAR\\
\raisebox{0.2em}[\height][\depth]{\fboxsep=0.2em \fboxrule=0.1em\fcolorbox{black}{white!40!white}{\null}}\quad FaceNet~\cite{facenet} & 99.67 && 23.34 && 99.70 && \textbf{99.96}\\
\crule{0.5em}{0.5em}\quad SphereFace~\cite{sphereface} & 97.22 && 29.49 && \textbf{99.34} && 98.71\\
\crule{0.5em}{0.5em}\quad ArcFace~\cite{arcface} & \textbf{64.53} && 03.43 && 33.25 && 35.30\\
\crule{0.5em}{0.5em}\quad COTS-A & \textbf{82.98} && 08.89 && 18.74 &&  32.48\\
\crule{0.5em}{0.5em}\quad COTS-B & \textbf{60.71} && 05.05 && 01.49 && 18.75\\
\midrule
Structural Similarity & \textbf{0.95 $\pm$ 0.01} && 0.82 $\pm$ 0.12 && 0.29 $\pm$ 0.06 &&  0.25 $\pm$ 0.06\\
\midrule
Computation Time (s) & \textbf{0.01} && 3.22  && 11.74 && 0.03\\
\bottomrule\\
\toprule \textbf{Impersonation Attack} & \multicolumn{1}{c}{Proposed \NAME} &
\phantom{abc} & \multicolumn{1}{c}{A$^3$GN~\cite{adv_attention}} &\phantom{abc} & \multicolumn{1}{c}{PGD~\cite{pgd}} & \phantom{abc} &\multicolumn{1}{c}{FGSM~\cite{fgsm}}\\
\midrule
Attack Success Rate (\%) @ 0.1\% FAR\\
\raisebox{0.2em}[\height][\depth]{\fboxsep=0.2em \fboxrule=0.1em\fcolorbox{black}{white!40!white}{\null}}\quad FaceNet~\cite{facenet} & 20.85 $\pm$ 0.40 && 05.99 $\pm$ 0.19 && \textbf{76.79 $\pm$ 0.26}  && 13.04 $\pm$ 0.12\\
\crule{0.5em}{0.5em}\quad SphereFace~\cite{sphereface} & \textbf{20.19 $\pm$ 0.27} && 07.94 $\pm$ 0.19 && 09.03 $\pm$ 0.39 && 02.34 $\pm$ 0.03\\
\crule{0.5em}{0.5em}\quad ArcFace~\cite{arcface} & \textbf{24.30 $\pm$ 0.44} && 17.14 $\pm$ 0.29 && 19.50 $\pm$ 1.95 && 08.34 $\pm$ 0.21\\
\crule{0.5em}{0.5em}\quad COTS-A & \textbf{20.75 $\pm$ 0.35} && 15.01 $\pm$ 0.30 && 01.76 $\pm$ 0.10 && 01.40 $\pm$ 0.08\\
\crule{0.5em}{0.5em}\quad COTS-B & \textbf{19.85 $\pm$ 0.28} && 10.23 $\pm$ 0.50 && 12.49 $\pm$ 0.24  && 04.67 $\pm$ 0.16 \\
\midrule
Structural Similarity & \textbf{0.92 $\pm$ 0.02} && 0.69 $\pm$ 0.04 && 0.77 $\pm$ 0.04 && 0.48 $\pm$ 0.75\\
\midrule
Computation Time (s) & \textbf{0.01} && 0.04 && 11.74 && 0.03\\
\bottomrule
\end{tabular}
\begin{tablenotes}
\item {\footnotesize \raisebox{0.3em}[\height][\depth]{\fboxsep=0.2em\fboxrule=0.1em\fcolorbox{black}{white!40!white}{\null}}\quad White-box matcher (used for training)\quad\quad \crule{0.5em}{0.5em}\quad Black-box matcher (never used in training)}
\end{tablenotes}
\end{threeparttable}}
\caption{Attack success rates and structural similarities between probe and gallery images for obfuscation and impersonation attacks. Attack rates for obfuscation comprises of 484,514 comparisons and the mean and standard deviation across 10-folds for impersonation reported. The mean and standard deviation of the structural similarities between adversarial and probe images along with the time taken to generate a single adversarial image (on a Quadro M6000 GPU) also reported.}
\label{tab:recognition}
\end{table*}

\noindent We apply the \texttt{tanh} activation function on the last convolution layer of the generator to ensure that the generated image $\in[-1, 1]$. In the paper, we denoted the output of the tanh layer as an ``adversarial mask'', $\mathcal{G}(x) \in [-1,1]$ and $x \in [-1,1]$. The final adversarial image is computed as $x_{adv} = 2 \times \texttt{clamp}\left[ \mathcal{G}(x) + \left(\frac{x+1}{2}\right) \right]_{0}^{1} -1$. This ensures $\mathcal{G}(x)$ can either add or subtract pixels from $x$ when $\mathcal{G}(x) \neq 0$. When $\mathcal{G}(x)\to0$, then $x_{adv}\to x$.

\paragraph{Face Recognition Systems} For all our experiments, we employ 5 state-of-the-art face matchers\footnote{All the open-source and COTS matchers achieve 99\% accuracy on LFW under LFW protocol.}. Three of them are publicly available, namely, FaceNet~\cite{facenet}, SphereFace~\cite{sphereface}, and ArcFace~\cite{arcface}. We also report our results on two commercial-off-the-shelf (COTS) face matchers, COTS-A and COTS-B\footnote{Both COTS-A and COTS-B utilize CNNs for face recognition. COTS-B is one of the top performers in the NIST Ongoing Face Recognition Vendor Test (FRVT)~\cite{nist_2018}.}. We use FaceNet~\cite{facenet} as the white-box face recognition model, $\mathcal{F}$, during training. \emph{All the testing images in this paper are generated from the same model (trained only with FaceNet) and tested on different matchers.}

\subsubsection{Comparison with Prevailing Adversarial Face Generators}
We compare our adversarial face synthesis method with state-of-the-art methods that have specifically been implemented or proposed for faces, including GFLM~\cite{gflm}, PGD~\cite{pgd}, FGSM~\cite{fgsm}, and A$^3$GN~\cite{adv_attention}\footnote{We train the baselines using their official implementations (detailed in the supplementary material).}. In Table~\ref{tab:recognition}, we find that compared to the state-of-the-art, \NAME generates adversarial faces that are similar to the probe~\ref{fig:comparison}.

Moreover, the adversarial images attain a high obfuscation attack success rate on 4 state-of-the-art black-box AFR systems in both obfuscation and impersonation settings. \NAME learns to perturb the salient regions of the face, unlike PGD~\cite{pgd} and FGSM~\cite{fgsm}, which alter every pixel in the image. GFLM~\cite{gflm}, on the other hand, geometrically warps the face images and thereby, results in low structural similarity. In addition, the state-of-the-art matchers are robust to such geometric deformation which explains the low success rate of GFLM on face matchers. A$^3$GN, another GAN-based method, however, fails to achieve a reasonable success rate in an impersonation setting.

\subsubsection{Ablation Study}
\begin{figure}[!h]
\setlength\tabcolsep{0px}
\newcolumntype{Y}{>{\centering\arraybackslash}X}
    \centering
    \captionsetup{font=footnotesize}
    \begin{tabularx}{\linewidth}{YYYYYY}
    \toprule
        Input & w/o $\mathcal{D}$  & w/o $\EL_{prt}$ & w/o $\EL_{idt}$ & with all  \\
    \midrule
        \includegraphics[width=0.97\linewidth]{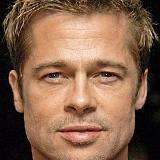} & 
        \includegraphics[width=0.97\linewidth]{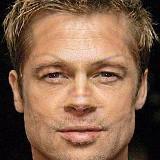} &
        \includegraphics[width=0.97\linewidth]{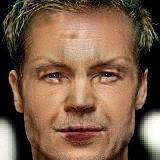} &
        \includegraphics[width=0.97\linewidth]{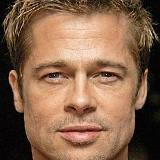} &
        \includegraphics[width=0.97\linewidth]{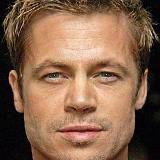} \\
        \includegraphics[width=0.97\linewidth]{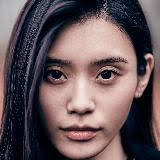} & 
        \includegraphics[width=0.97\linewidth]{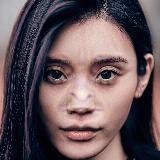} &
        \includegraphics[width=0.97\linewidth]{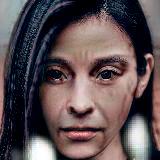} &
        \includegraphics[width=0.97\linewidth]{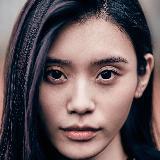} &
        \includegraphics[width=0.97\linewidth]{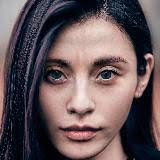} \\
    \bottomrule
    \end{tabularx}
    \caption{Variants of \NAME trained without the discriminator, perturbation loss, and identity loss, respectively. Every component of \NAME is necessary.}
    \label{fig:ablation}
\end{figure}

In order to analyze the importance of each module in our system, in Figure~\ref{fig:ablation}, we train three variants of \NAME for comparison by removing the discriminator ($\mathcal{D}$), perturbation loss $\EL_{perturbation}$, and identity loss $\EL_{identity}$, respectively. 

 The discriminator helps to ensure the visual quality of the synthesized faces are maintained. With the generator alone, undesirable artifacts are introduced. Without the proposed perturbation loss, perturbations in the adversarial mask are unbounded and therefore, leads to a lack in perceptual quality. The identity loss is imperative in ensuring an adversarial image is obtained. Without the identity loss, the synthesized image cannot evade state-of-the-art face matchers. 
We find that every component of \NAME is necessary in order to obtain an adversarial face that is not only perceptually realistic but can also evade state-of-the-art face matchers.

\subsubsection{What is \NAME Learning?}
\begin{figure}[!h]
\captionsetup{font=footnotesize}
\footnotesize
\centering\begin{tabular}{@{}c@{ }c@{ }c@{ }c@{ }c@{}}
\textbf{Probe} & \textbf{Adv. Mask} & \textbf{Visualization} & \textbf{Adv. Image} \\
\includegraphics[width=.2\textwidth]{images/fig/3_o.jpg}&
\includegraphics[width=.2\textwidth]{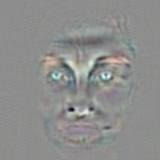}&
\includegraphics[width=.2\textwidth]{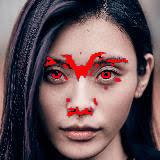}&
\includegraphics[width=.2\textwidth]{images/fig/3_g.jpg}\\
& & &\footnotesize 0.12\\
\includegraphics[width=.2\textwidth]{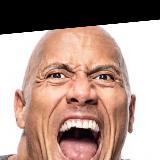}&
\includegraphics[width=.2\textwidth]{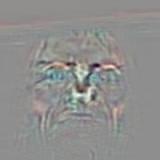}&
\includegraphics[width=.2\textwidth]{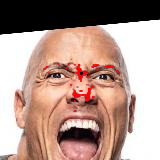}&
\includegraphics[width=.2\textwidth]{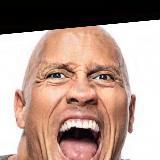}\\
& & & {\footnotesize 0.26}%
\end{tabular}
\caption{State-of-the-art face matchers can be evaded by slightly perturbing salient facial regions, such as eyebrows, eyeballs, and nose (cosine similarity obtained via ArcFace~\cite{arcface}).}%
\label{fig:visualization}
\end{figure}

Via $\EL_{perturbation}$, during training, \NAME learns to perturb only the salient facial regions that can evade the face matcher, $\mathcal{F}$ (FaceNet~\cite{facenet} in our case). In Figure~\ref{fig:visualization}, \NAME synthesizes the adversarial masks corresponding to the probes. We then threshold the mask to extract pixels with perturbation magnitudes exceeding $0.40$. It can be inferred that the eyebrows, eyeballs, and nose contain highly discriminative information that an AFR system utilizes to identify an individual. Therefore, perturbing these salient regions are enough to evade state-of-the-art face recognition systems.

\subsubsection{Transferability of \NAME}
\begin{figure}[!h]
    \centering
    \captionsetup{font=footnotesize}
    \includegraphics[width=0.8\linewidth]{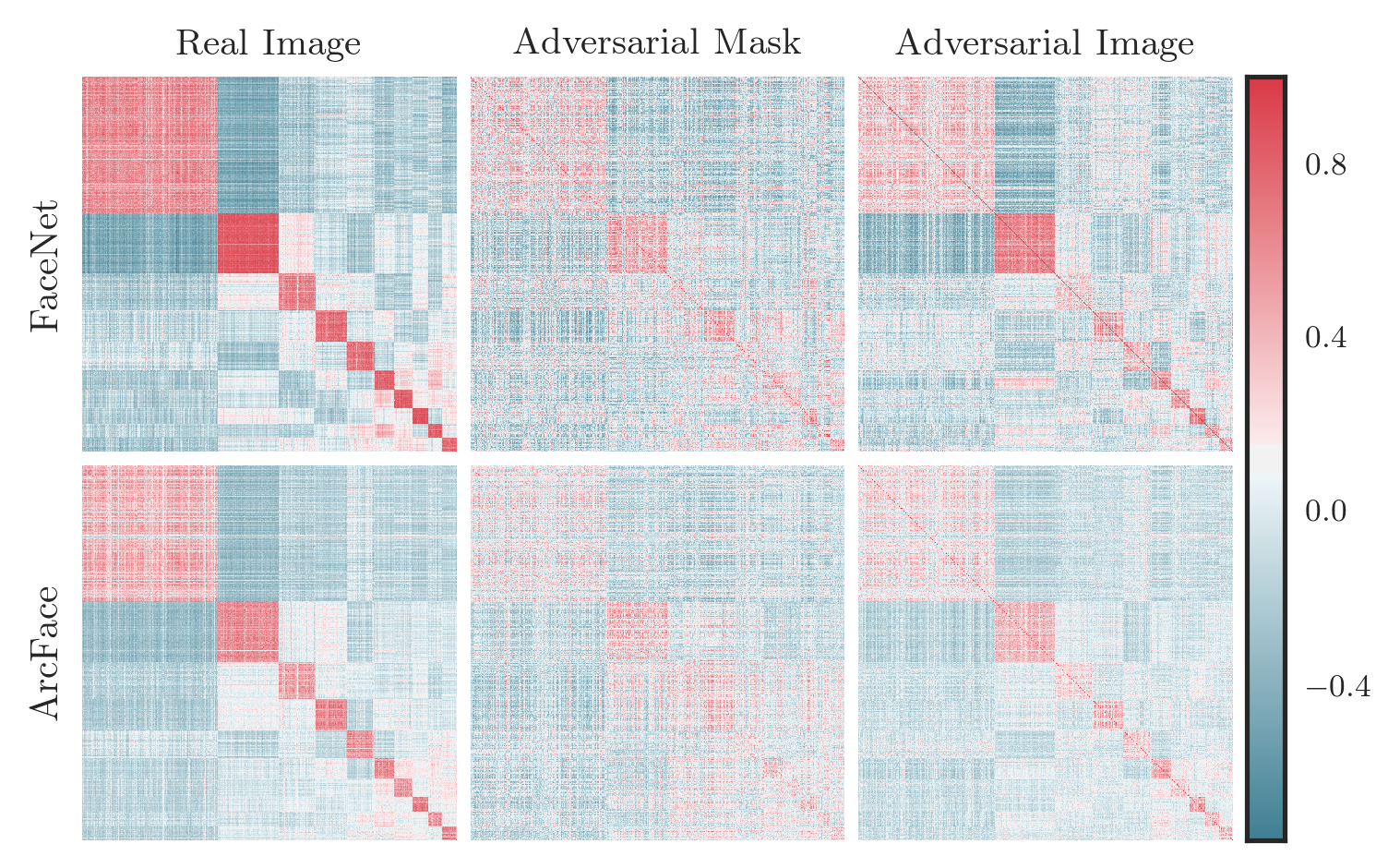}
    \vspace{-1.0em}\caption{Correlation between face features extracted via FaceNet and ArcFace from 1,456 images belonging to 10 subjects.}
    \label{fig:corr}
\end{figure}

In Table~\ref{tab:recognition}, we find that attacks synthesized by \NAME when trained on a white-box matcher (FaceNet), can successfully evade 5 other face matchers that are not utilized during training in both obfuscation and impersonation settings. In order to investigate the transferability property of \NAMENOSPACE, we extract face embeddings of real images and their corresponding adversarial images, under the obfuscation setting, via the white-box matcher (FaceNet) and a black-box matcher (ArcFace). In total, we extract feature vectors from 1,456 face images of 10 subjects in the LFW dataset~\cite{lfw}. In Figure~\ref{fig:corr}, we plot the correlation heatmap between face features of real images, their corresponding adversarial masks and adversarial images. First, we observe that face embeddings of real images extracted by FaceNet and ArcFace are correlated in a similar fashion. This indicates that both matchers extract features with related pairwise correlations. Consequently, perturbing salient features for FaceNet can lead to high attack success rates for ArcFace as well. The similarity among the correlation distributions of both matchers can also be observed when adversarial masks and adversarial images are input to the matchers. That is, receptive fields for automatic face recognition systems attend to similar regions in the face. 

\begin{figure}[!h]
    \centering
    \captionsetup{font=footnotesize}
    \includegraphics[width=0.9\linewidth]{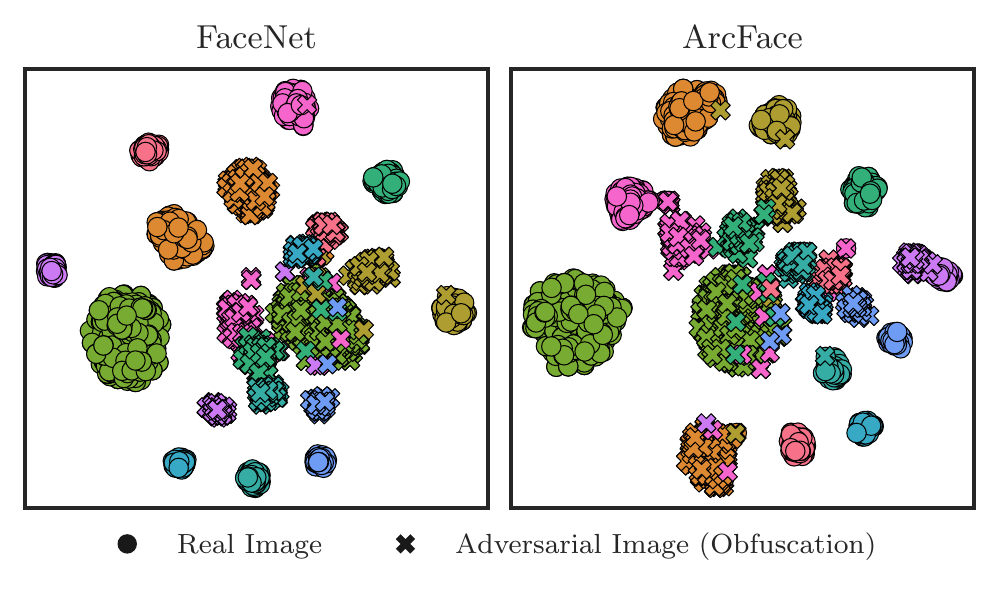}
    \vspace{-1.0em}\caption{2D t-SNE visualization of face representations extracted via FaceNet and ArcFace from 1,456 images belonging to 10 subjects.}
    \label{fig:tsne}
\end{figure}

To further illustrate the distributions of the embeddings of real and synthesized images, we plot the 2D t-SNE visualization of the face embeddings for the 10 subjects in Figure~\ref{fig:tsne}. The identity clusterings can be clearly observed from both real and adversarial images. In particular, the adversarial counterpart of each subject forms a new cluster that draws closer to the adversarial clusterings of other subjects. This shows that \NAME perturbs only salient pixels related to face identity while maintaining a semantic meaning in the feature space, resulting in a similar manifold of synthesized faces to that of real faces.

\subsubsection{Controllable Perturbation}
\begin{figure}[!h]
    \centering
    \captionsetup{font=footnotesize}
    \includegraphics[width=0.7\linewidth]{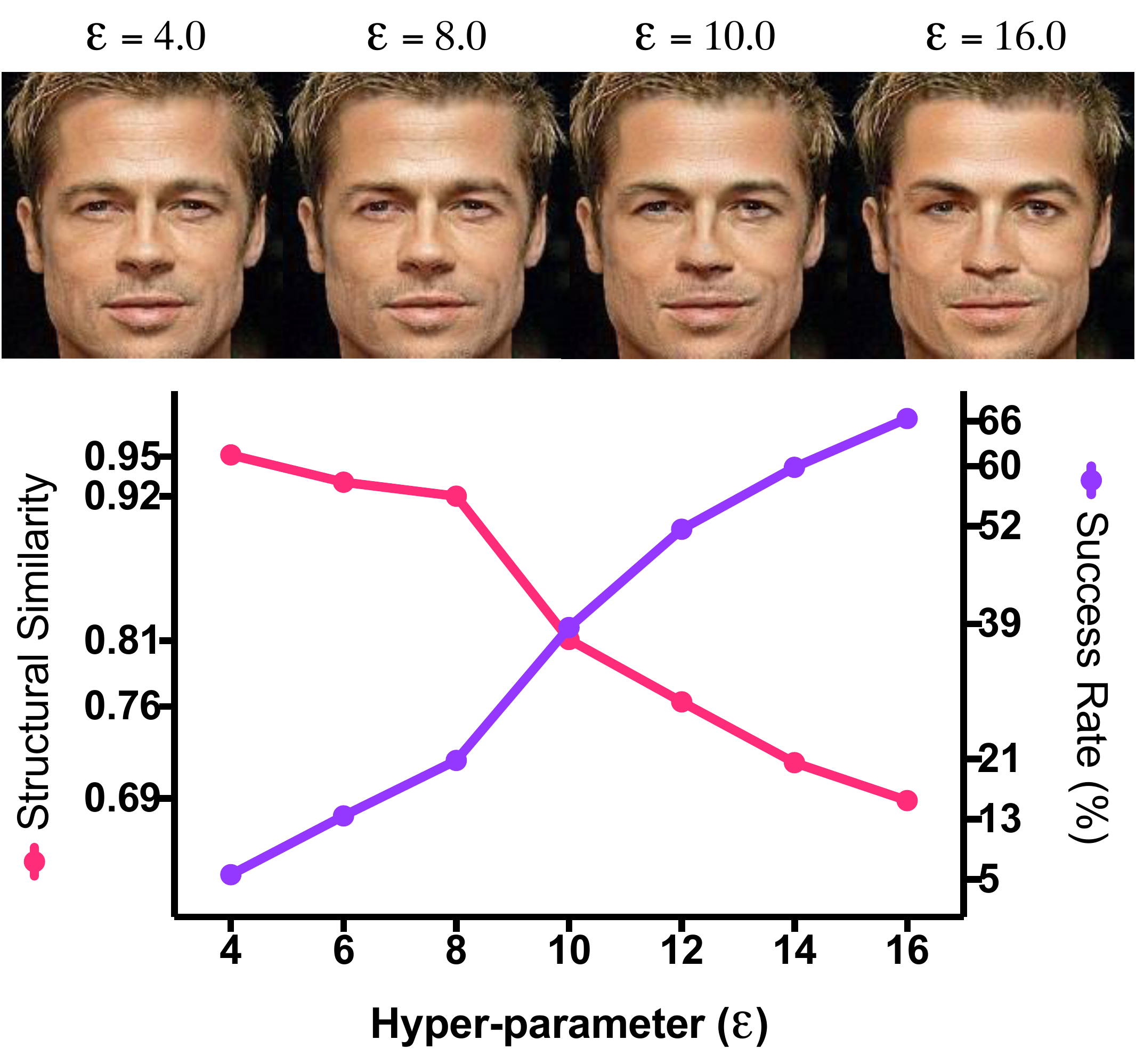}
    \caption{Trade-off between attack success rate and structural similarity for impersonation attacks.}
    \label{fig:min_eps}
\end{figure}

The perturbation loss, $\EL_{perturbation}$ is bounded by a hyper-parameter, $\epsilon$,~\ie, the $L_2$ norm of the adversarial mask must be at least $\epsilon$. Without this constraint, the adversarial mask becomes a blank image with no changes to the probe. With $\epsilon$, we can observe a trade-off between the attack success rate and the structural similarity between the probe and synthesized adversarial face (Fig.~\ref{fig:min_eps}). A higher $\epsilon$ leads to less perturbation restriction, resulting in a higher attack success rate at the cost of a lower structural similarity. For an impersonation attack, this implies that the adversarial image may contain facial features from both the hacker and the target. In our experiments, we chose $\epsilon = 8.0$ and $\epsilon = 3.0$ for impersonation and obfuscation attacks, respectively.

\subsubsection{Attacks via \NAME Beyond Faces}

We now show that the \NAMENOSPACE method, coupled with the proposed Minutiae Displacement and Distortion Modules, can be extended to effectively generate adversarial fingerprints which are visually similar to corresponding probe fingerprints while evading two state-of-the-art COTS fingerprint matchers as well as a deep network-based fingerprint matcher.

\begin{figure*}[!h]
\vspace{-1.2em}
    \centering
    \footnotesize
    \captionsetup{font=footnotesize}
    \setlength{\fboxsep}{2.2pt}
    \setlength{\fboxrule}{0.2pt}
    \begin{minipage}{0.189\linewidth}
    \fcolorbox{white}{white}{\includegraphics[width=0.95\linewidth]{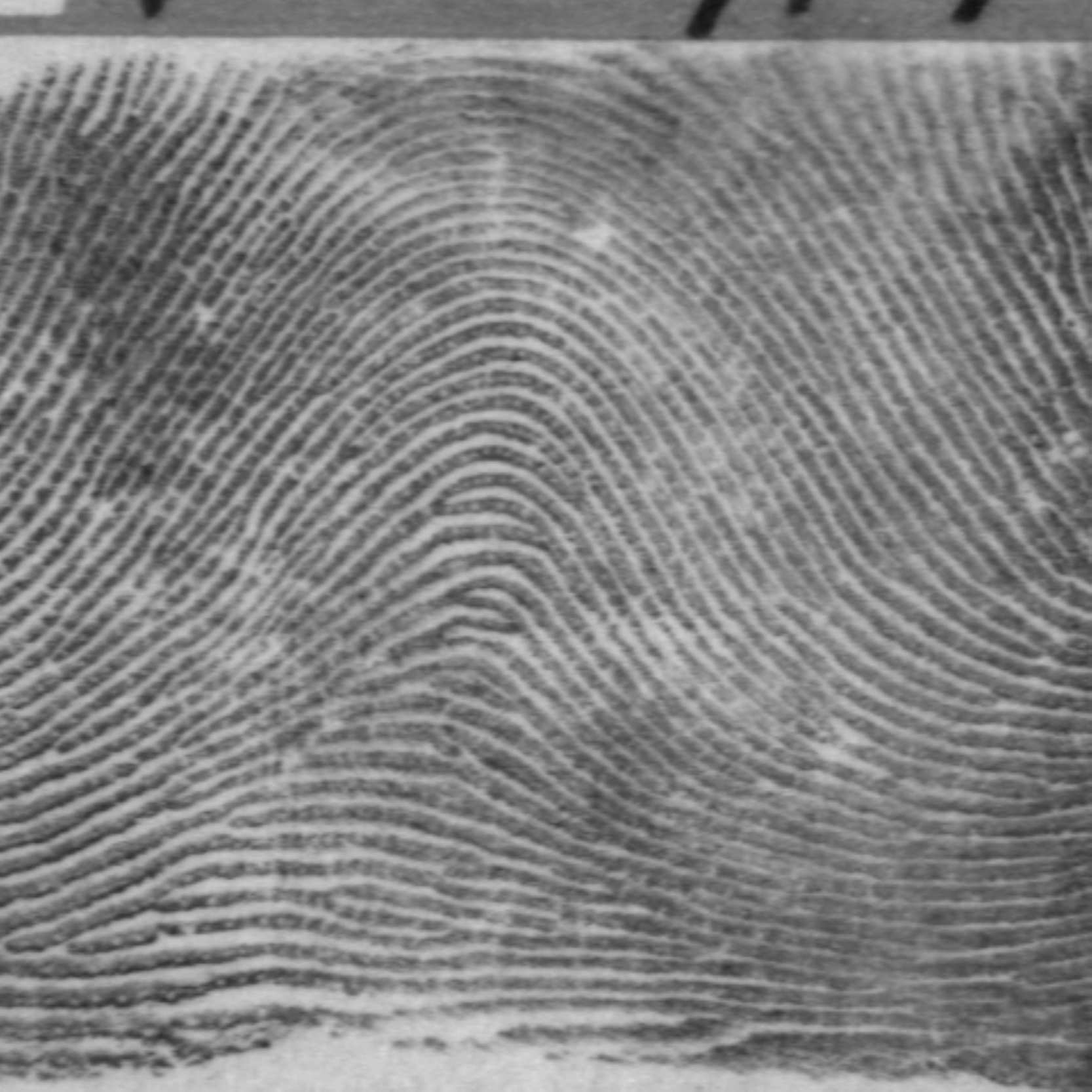}}\\
    \centering{\small{FS: 0.97\vskip 0.1em}}
    \fcolorbox{white}{white}{\includegraphics[width=0.95\linewidth]{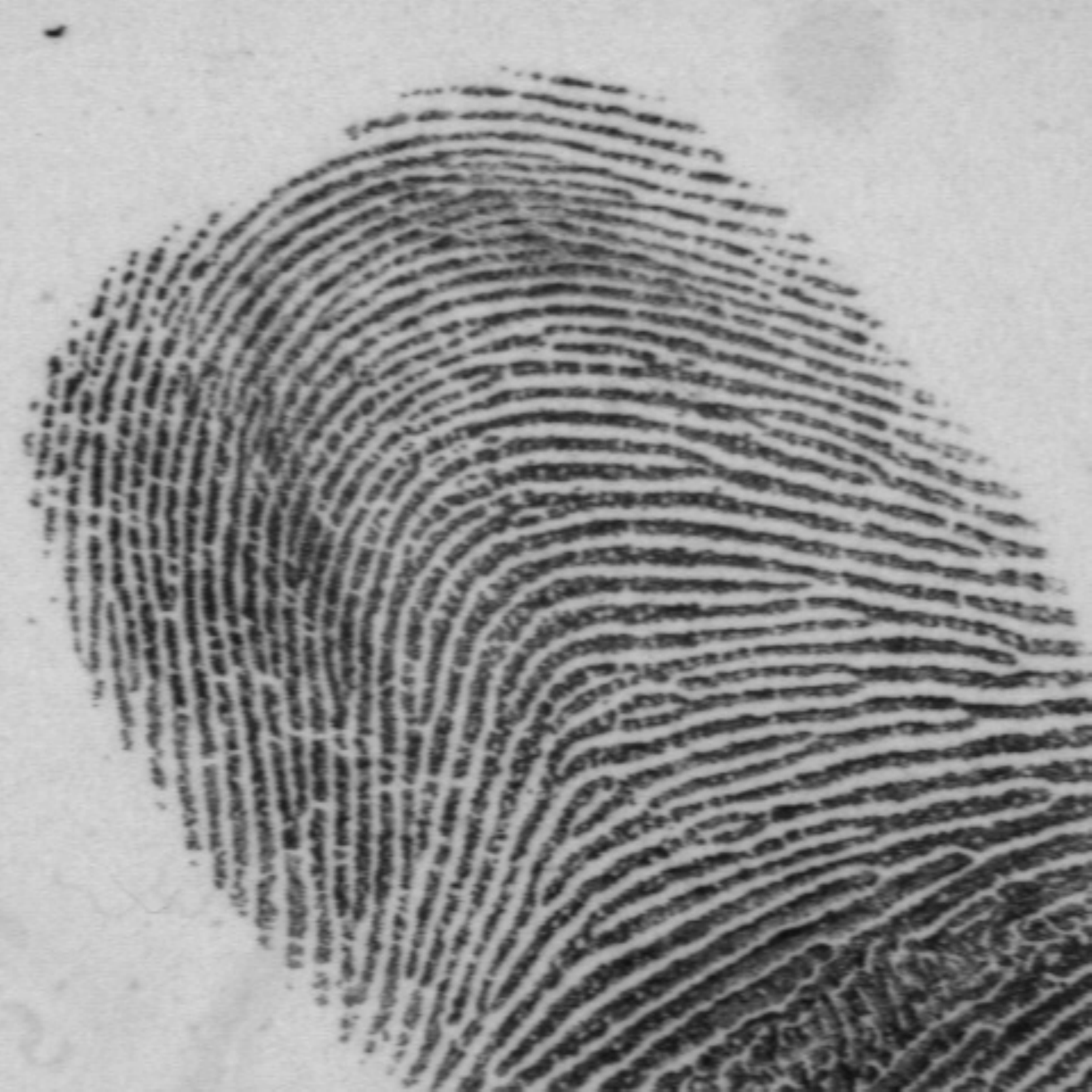}}\\
    \centering{\small{FS: 0.92\vskip 0.5em (a) Enrolled Mate}}
    \end{minipage}\;
    \begin{minipage}{0.189\linewidth}
    \fcolorbox{white}{white}{\includegraphics[width=0.95\linewidth]{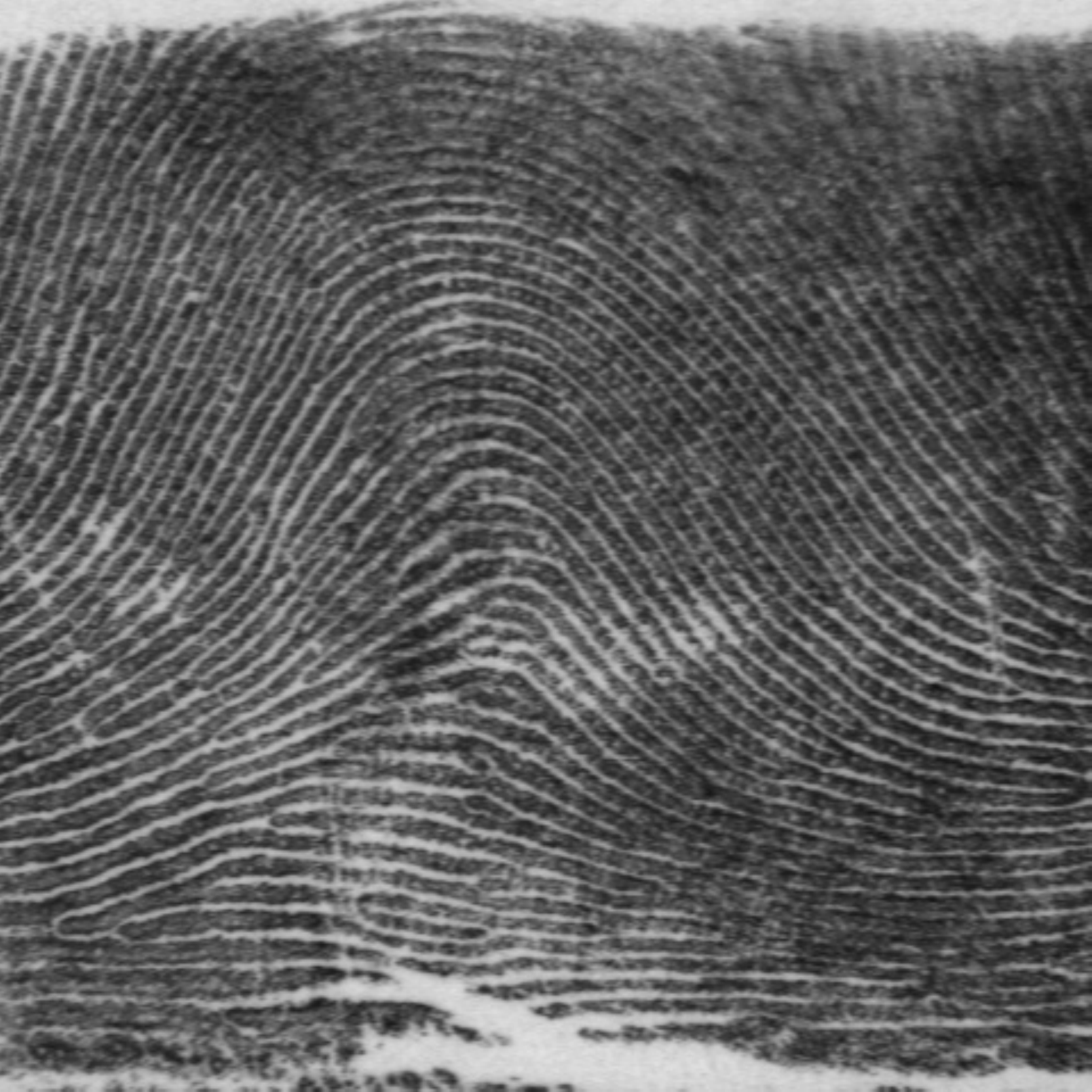}}\\
    \centering {\small{VS: 235 $|$ FS: 0.96\vskip 0.1em}}
    \fcolorbox{white}{white}{\includegraphics[width=0.95\linewidth]{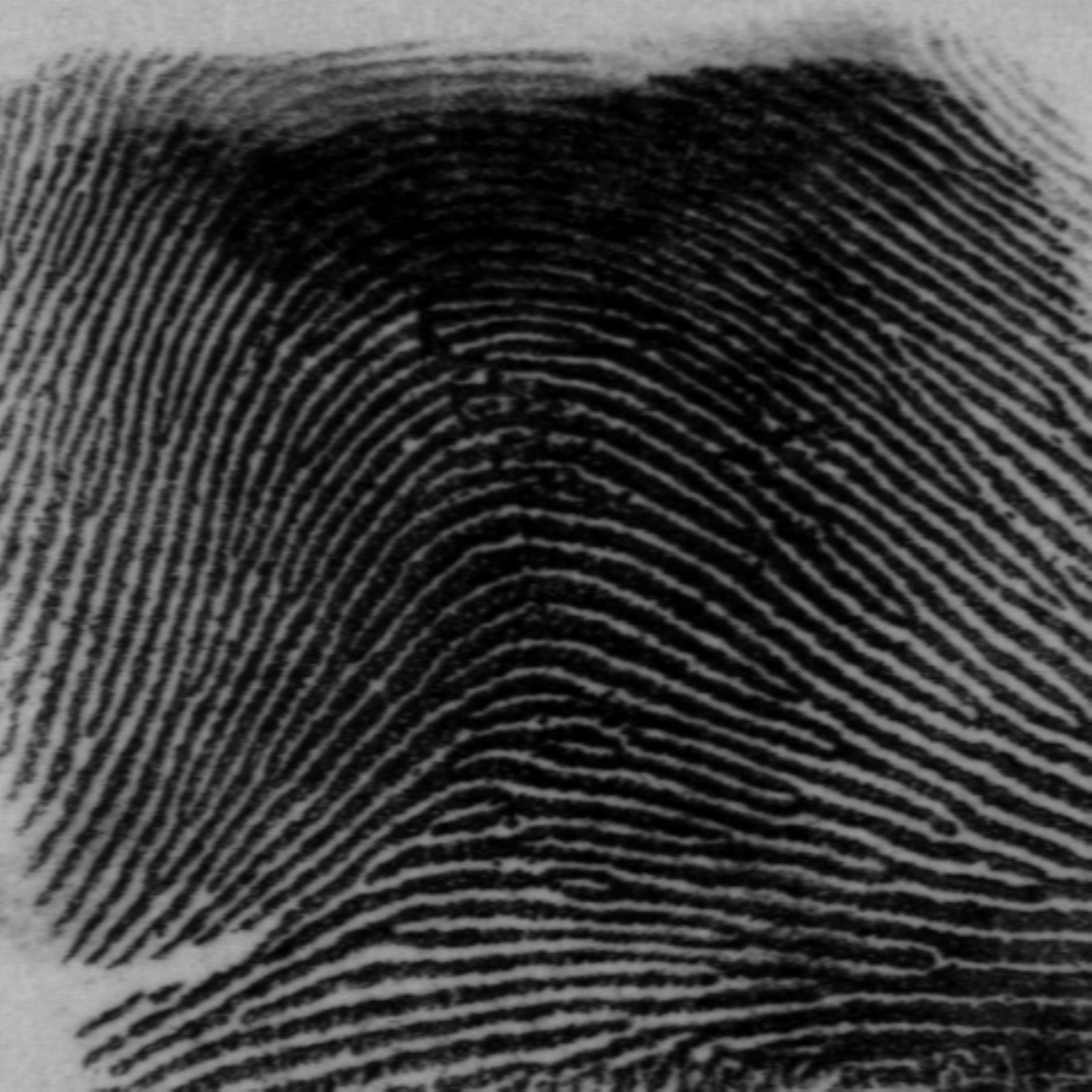}}\\
    \centering {\small{VS: 172 $|$ FS: 0.99\vskip 0.5em (b) Input Probe}}
    \end{minipage}\;
    \begin{minipage}{0.189\linewidth}
    \fcolorbox{white}{white}{\includegraphics[width=0.95\linewidth]{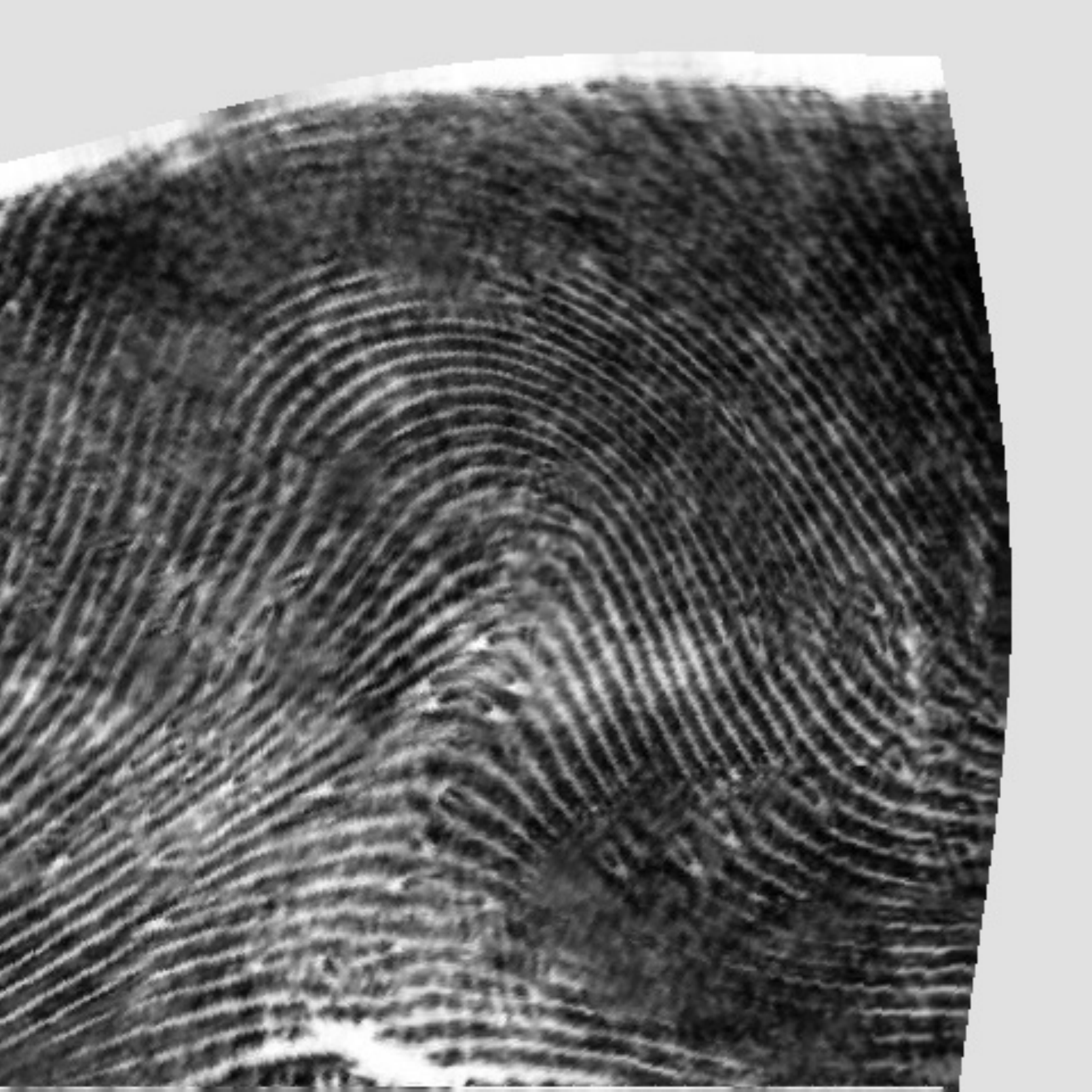}}\\
     \centering{\small{VS: 31 $|$ FS: 0.92\vskip 0.1em}}
    \fcolorbox{white}{white}{\includegraphics[width=0.95\linewidth]{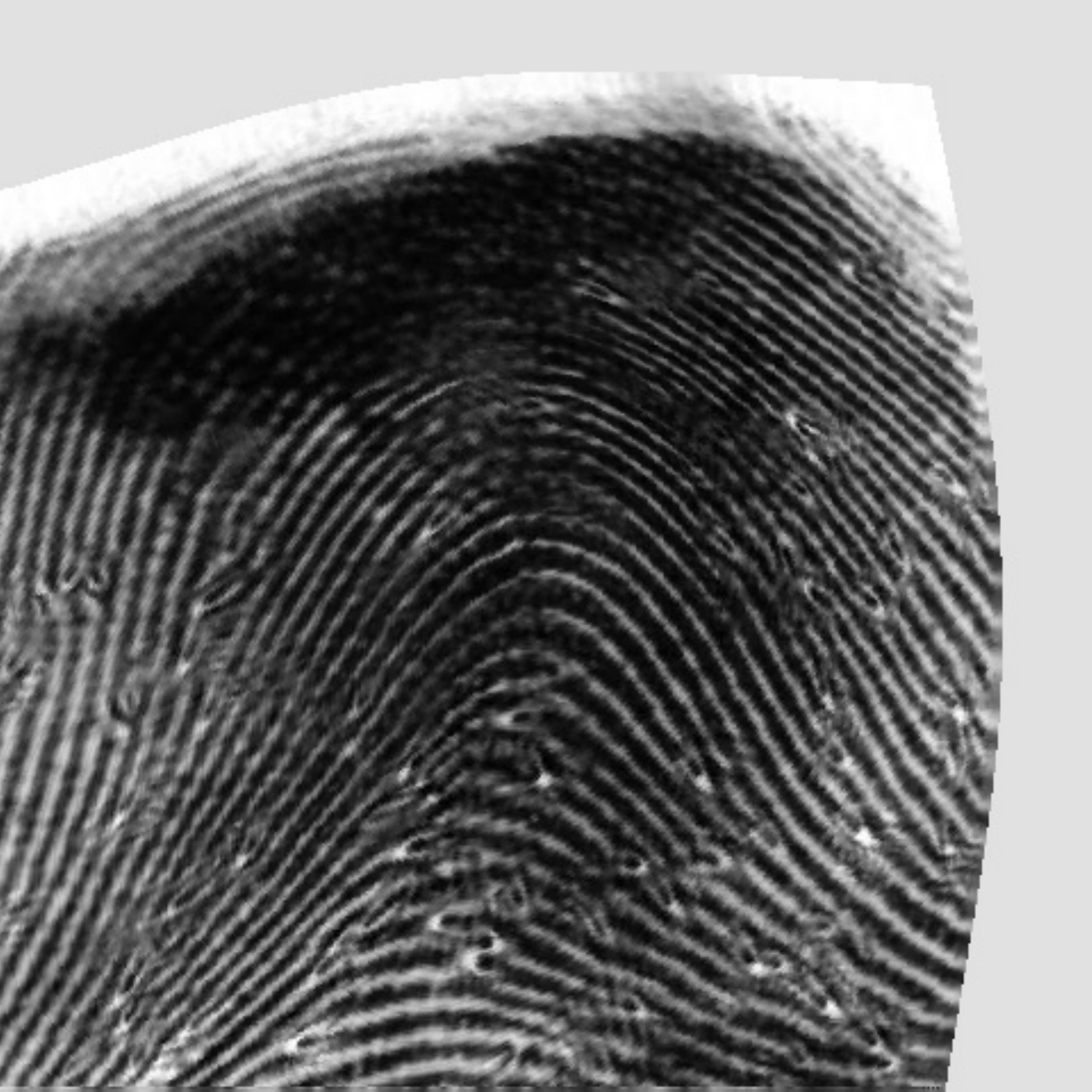}}\\
    \centering {\small{VS: 10 $|$ FS: 0.92\vskip 0.5em (c) \NAME}}
    \end{minipage}\;
    \begin{minipage}{0.189\linewidth}
    \fcolorbox{white}{white}{\includegraphics[width=0.95\linewidth]{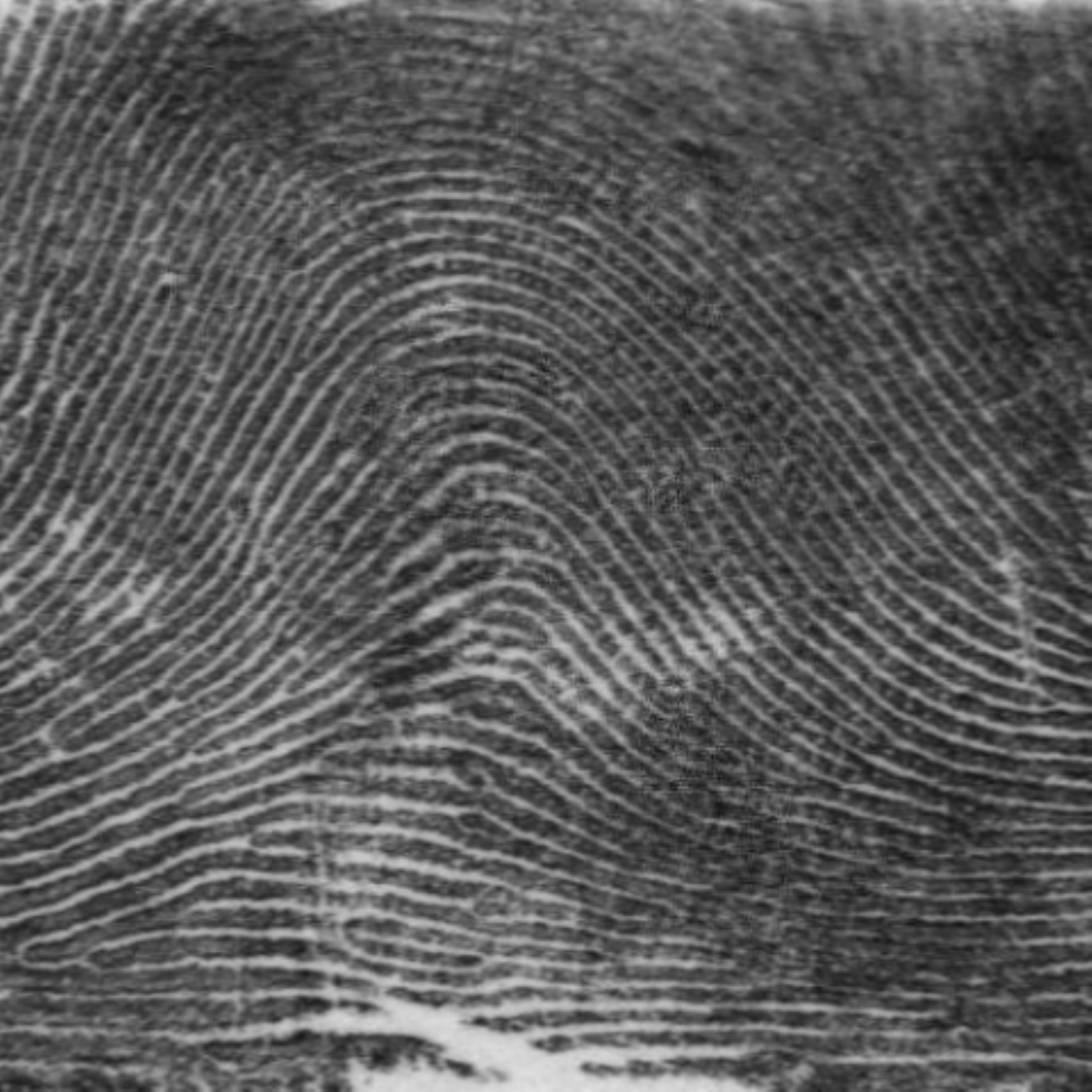}}\\
    \centering{\small{VS: 134 $|$ FS: 0.96\vskip 0.1em}}
    \fcolorbox{white}{white}{\includegraphics[width=0.95\linewidth]{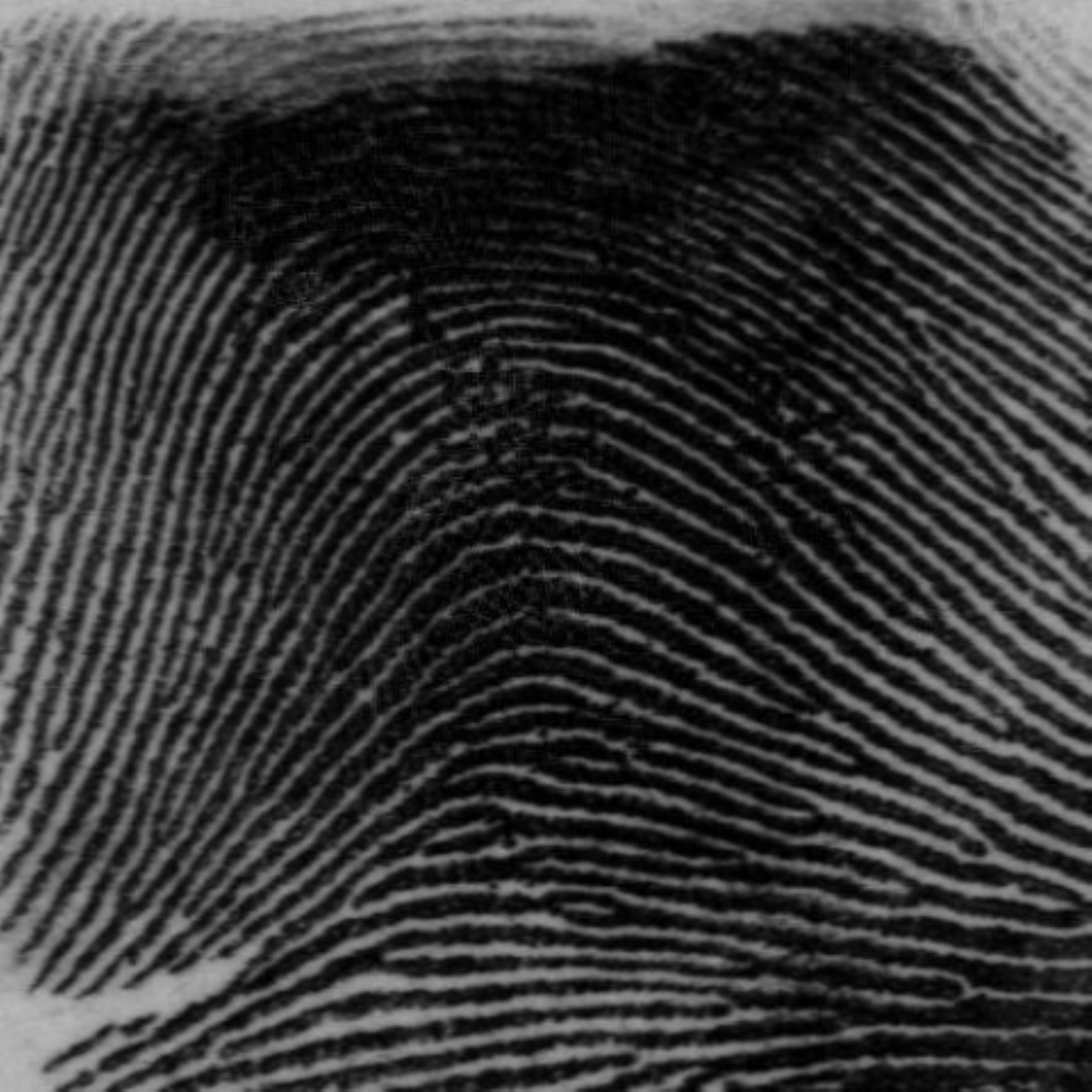}}\\
    \centering {\small{VS: 104 $|$ FS: 0.96\vskip 0.5em (d) DeepFool~\cite{deepfool}}}
    \end{minipage}\;
    \begin{minipage}{0.189\linewidth}
    \fcolorbox{white}{white}{\includegraphics[width=0.95\linewidth]{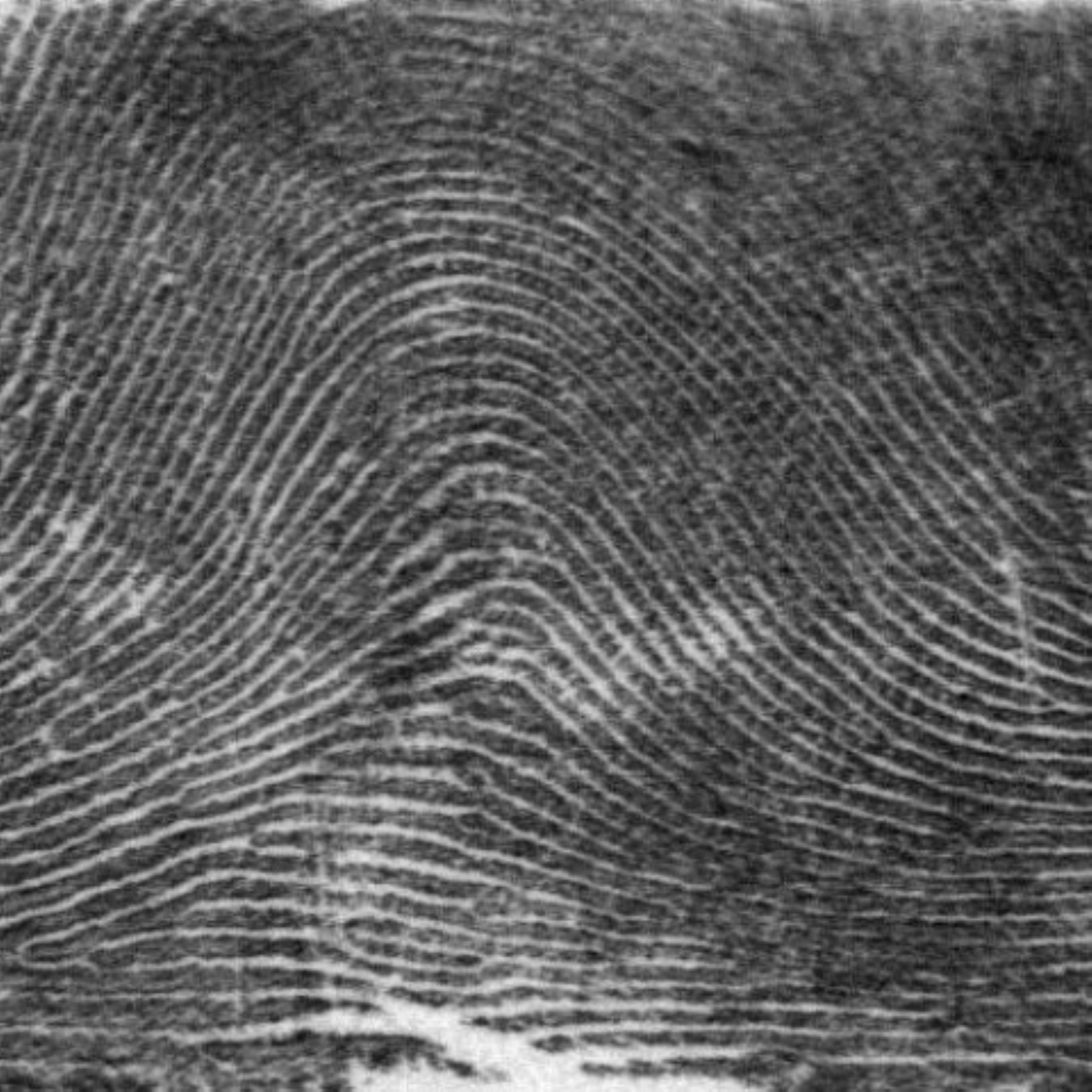}}\\
    \centering{\small{VS: 139 $|$ FS: 0.95\vskip 0.1em}}
    \fcolorbox{white}{white}{\includegraphics[width=0.95\linewidth]{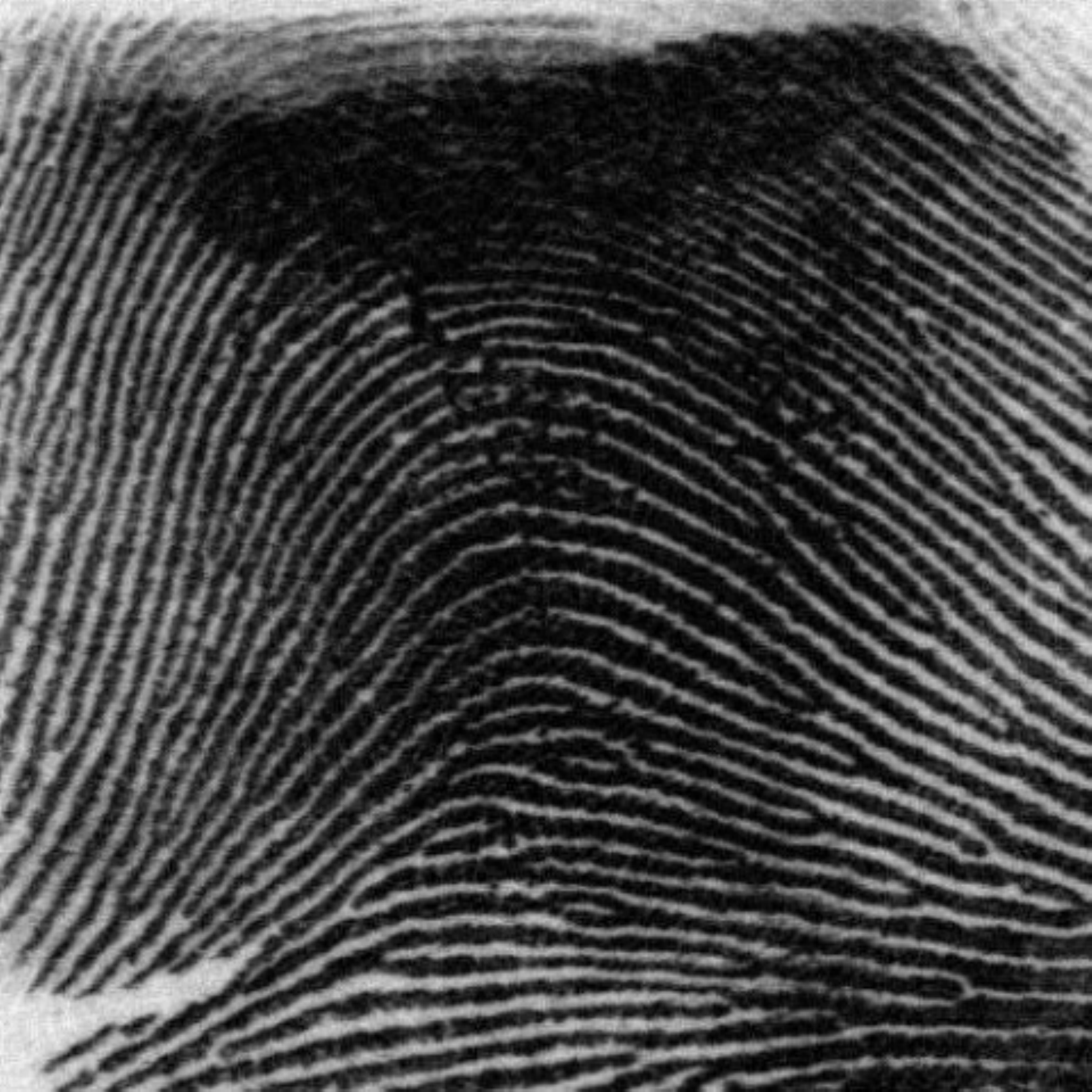}}\\
    \centering {\small{VS: 104 $|$ FS: 0.96\vskip 0.5em (d) PGD~\cite{pgd}}}
    \end{minipage}\;
    \captionof{figure}{\small{Example probe and corresponding mate fingerprints along with synthesized adversarial probes. (a) Two example mate fingerprints from NIST SD4~\cite{nist_sd4}, and (b) the corresponding mates. Adversarial probe fingerprints using different approaches are shown in: (c) proposed synthesis method, \NAMENOSPACE; (d-e) state-of-the-art methods, DeepFool and PGD respectively. VeriFinger v11.0 match score (probe v. mate) - VS, and the fingerprintness score (degree of similarity of a given image to a fingerprint pattern) - FS $\in$ [0,1]~\cite{fingerprintness}, which ranges from 1 (the highest) to 0 (the lowest), are given below each image. A VS of above 48 (at 0.01\% FAR) indicates a successful match between the probe and the mate. The proposed attack \NAME successfully evades COTS and deep network-based matchers, while maintaining visual fingerprint perceptibility and high fingerprintness scores.}}
    \label{fig:example_attack}
\end{figure*}

Grosz \textit{et. al}~\cite{whitebox} showed that random minutiae position displacements and non-linear distortions drastically affected the performance of COTS fingerprint matchers. \NAME builds upon these two perturbations and when given a probe fingerprint, can synthesize an adversarial fingerprint image that retains all of the original fingerprint attributes except the identity, i.e. a fingerprint recognition system should not match the adversarial fingerprint to the probe fingerprint (obfuscation attack).

\begin{figure*}[!h]
  \includegraphics[width=\linewidth]{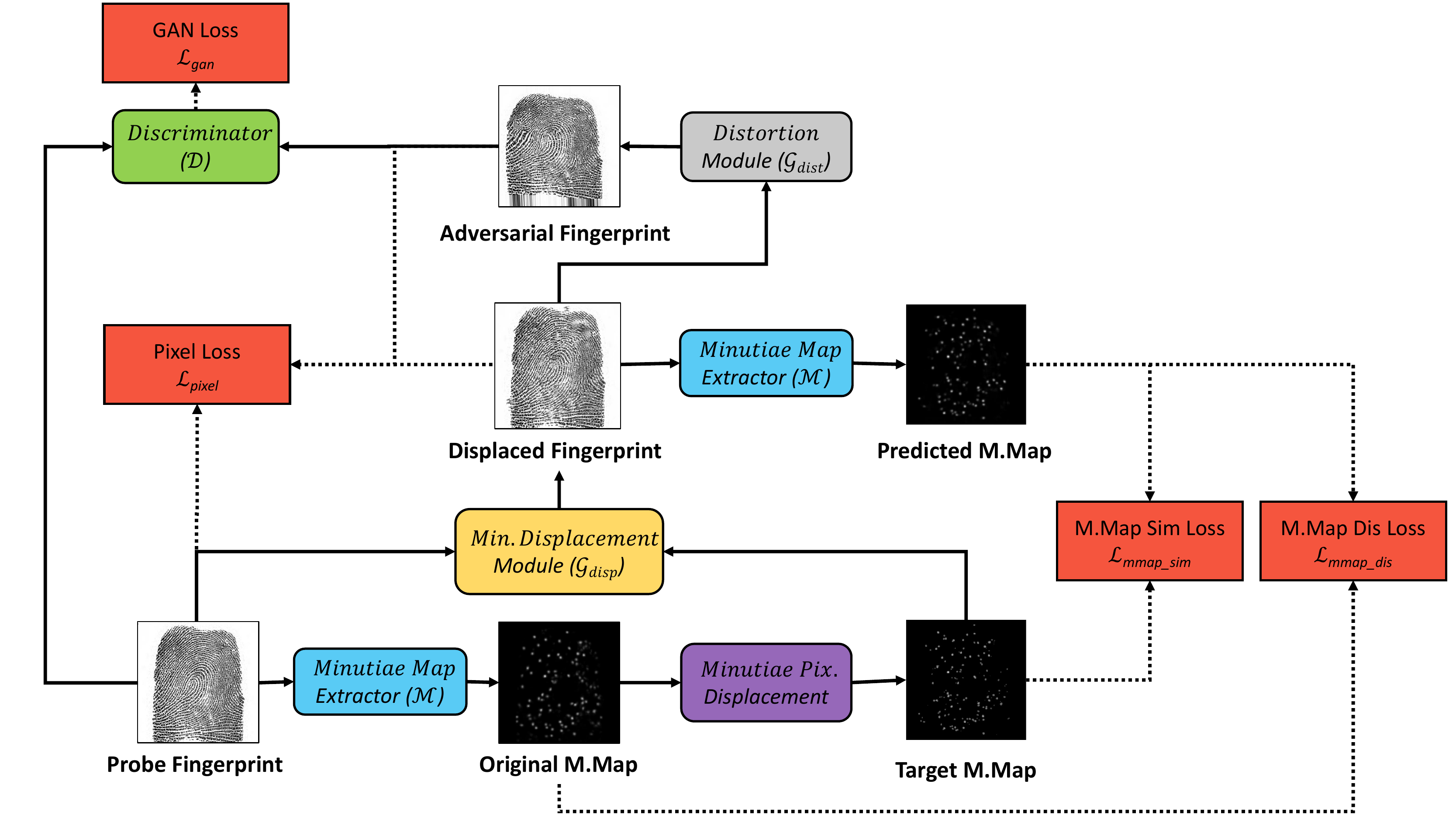}
  \captionsetup{font=footnotesize}
  \caption{Schematic of \NAME for generating adversarial fingerprints. Given a probe fingerprint image, it is passed to $\mathcal{G}_{disp}$ which randomly displaces its minutiae points. The distortion module ($\mathcal{G}_{dist}$) identifies control points on the displaced fingerprint and non-linearly distorts the image to output the adversarial fingerprint. The solid black arrows show the forward pass of the network while the dotted black arrows show the propagation of the losses.}
  \label{fig:AdvFinge_Architecture}
\end{figure*}

Figure \ref{fig:AdvFinge_Architecture} shows the schematic of \NAME conditioned for fingerprints. The following subsections explain the major components of the approach in detail.

\paragraph{Minutiae Displacement Module} While the authors in ~\cite{whitebox} showed the effectiveness of random minutiae position displacements on COTS matchers, they studied the effect of this perturbation by directly modifying the minutiae template instead of the fingerprint image (pixel space). However, it may be difficult to obtain the minutiae template of a given fingerprint image using COTS minutiae extractors rather than the source fingerprint image itself. Thus, we propose a minutiae displacement module $\mathcal{G}_{disp}$ which, given a fingerprint image, displaces its minutiae points in random directions by a predefined distance. To extract minutiae points from a fingerprint image, we employ a minutiae map extractor ($\mathcal{M}$) from \cite{mmap}. For a fingerprint image of width \textit{w} and height \textit{h}, $\mathcal{M}$ outputs a 12 channel heat map $\textit{H} \in \mathbb{R}^{\textit{h} \times \textit{w} \times 12}$, where if \textit{H}(\textit{i},\textit{j},\textit{c}), value of the heat map at position (\textit{i},\textit{j}) and channel \textit{c}, is greater than a threshold $\textit{m}_{t}$ and is the local maximum in its $5 \times 5 \times 3$ neighboring cube, a minutiae is marked at (\textit{i},\textit{j}). The minutiae direction $\theta$ is calculated by maximising the quadratic interpolation with respect to: 

\begin{gather}
    f\left(\left(c-1\right) \times \frac{\pi}{6}\right) = \textit{H}\left(\textit{i},\textit{j},(\textit{c}-1)\%12\right)\\
    f\left(c \times \frac{\pi}{6}\right) = \textit{H}(\textit{i},\textit{j},\textit{c})\\
    f\left((c+1) \times \frac{\pi}{6}\right) = \textit{H}(\textit{i},\textit{j},(\textit{c}+1)\%12)
\end{gather}

\begin{figure*}[!h]
  \includegraphics[width=\linewidth]{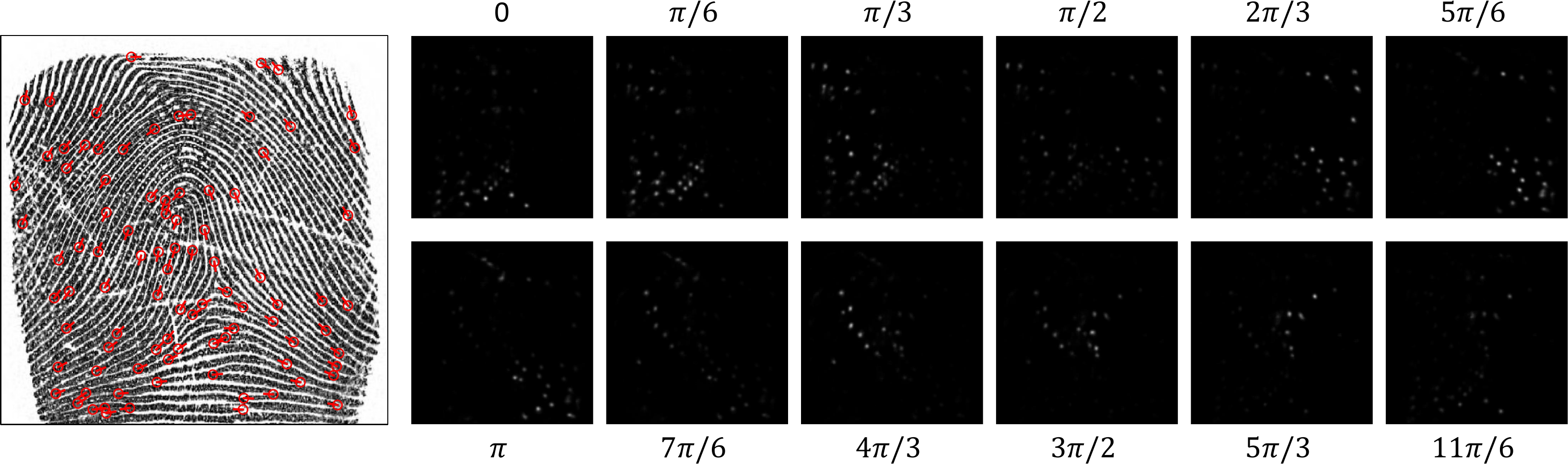}
  \captionsetup{font=footnotesize}
  \caption{The 12 channel minutiae map of an example fingerprint image shown on the left. The minutiae points (shown in red) are marked by a COTS minutiae extractor. The bright spots in each channel image indicate the spatial location of minutiae points while the $\textit{k}^{th}$ channel ($\textit{k} \in [0,11]$) indicate the contributions of minutiae points to the $\textit{k}\pi/6$ orientation.}
  \label{fig:mmap}
\end{figure*}

Figure \ref{fig:mmap} shows a fingerprint image and its corresponding 12 channel minutiae map. Once $\mathcal{M}$ extracts a minutiae map $\textit{H}_{probe}$ from the input probe fingerprint $x$, we detect minutiae points by applying a threshold of 0.2 on $\textit{H}_{probe}$ and finding closed contours. Each detected contour, at say location (\textit{i}, \textit{j}), is displaced by a predefined $\textit{L}_{1}$ distance $d = |\Delta\textit{i}| + |\Delta\textit{j}|$, giving us the target minutiae map $\textit{H}_{target}$. 

\begin{figure}
  \includegraphics[width=\linewidth]{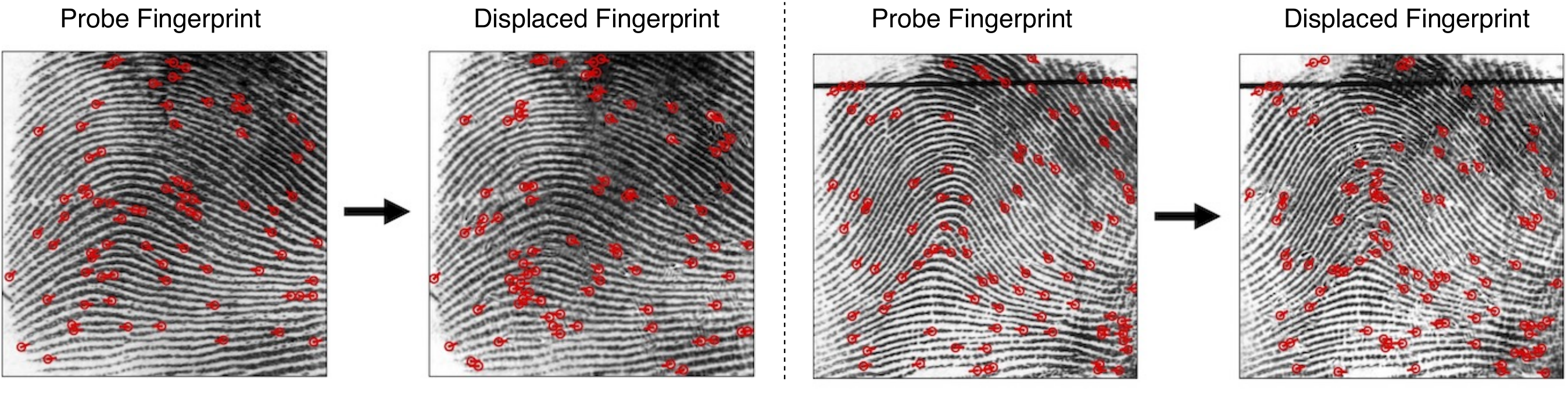}
  \captionsetup{font=footnotesize}
  \caption{Example probe fingerprints from NIST SD4~\cite{nist_sd4} and their corresponding output from the minutiae displacement module $\mathcal{G}_{disp}$. The minutiae points (shown in red) are marked using a COTS minutiae extractor.}
  \label{fig:disp}
\end{figure}

The minutiae displacement module $\mathcal{G}_{disp}$ is essentially an autoencoder conditioned on the probe fingerprint $x$ and the target minutiae map $\textit{H}_{target}$. It learns to generate a displaced fingerprint $x_{disp}$ whose predicted minutiae map $\textit{H}_{pred}$ is as close as possible to the target minutiae map $\textit{H}_{target}$ in the pixel space. To achieve this, we have three losses that govern $\mathcal{G}_{disp}$:

\begin{gather}
    \mathcal{L}_{mmap\_sim} = ||\textit{H}_{target} - \textit{H}_{pred}||_{1}\\
    \mathcal{L}_{mmap\_dis} = \frac{1}{||\textit{H}_{target} - \textit{H}_{probe}||_{1}}
\end{gather}

, where $\mathcal{L}_{mmap\_sim}$ is the minutiae map similarity loss which minimises the distance between the predicted and target minutiae map, while the minutiae map dissimilarity loss $\mathcal{L}_{mmap\_dis}$ maximises the distance between the predicted and probe minutiae map. In figure \ref{fig:disp}, we show two example probe fingerprints and their corresponding displaced fingerprints after passing through $\mathcal{G}_{disp}$.

\paragraph{Distortion Module}

One of the most noteworthy conclusions from \cite{whitebox} was that non-linear distortions to minutiae points was one of the most successful perturbations to lower the similarity scores between perturbed and corresponding unperturbed fingerprints. Again, the non-linear distortion was applied to all the minutiae points in the template and not to the image. Thus, our next step in generating adversarial fingerprints consists of a distortion module $\mathcal{G}_{dist}$ which learns to distort salient points in a fingerprint image.

The architecture of $\mathcal{G}_{dist}$ consists of an encoder conditioned on the input probe fingerprint $x$ and the target minutiae map $\textit{H}_{probe}$. The output from the encoder is a predefined number of control points\footnote{Control points are points in an image to which non-linear distortion is applied.} $c$. The non-linear distortion model proposed in~\cite{nonlineardistortion}, learned using a thin plate spline (TPS) model~\cite{tps} from 320 already distorted fingerprint videos, was employed to calculate the displacements of the predicted control points. The hyper-parameter $\sigma$ is used to indicate the extent of the distortion. The control points and their displacements are then fed to a differentiable warping module~\cite{warp} to get the resultant adversarial fingerprint $x_{adv}$.

To limit the magnitude of non-linear distortion and to ensure that $x_{disp}$ and $x_{adv}$ are close to the probe fingerprint $x$, we introduce pixel loss between the image pairs ($x$, $x_{disp}$) and ($x$, $x_{adv}$):

\begin{gather}
    \mathcal{L}_{pixel} = \frac{1}{n}\sum_{i,j}|x_{i,j} - x_{disp_{i,j}}| + \frac{1}{n}\sum_{i,j}|x_{i,j} - x_{adv_{i,j}}|
\end{gather}

\begin{figure}[!h]
  \includegraphics[width=\linewidth]{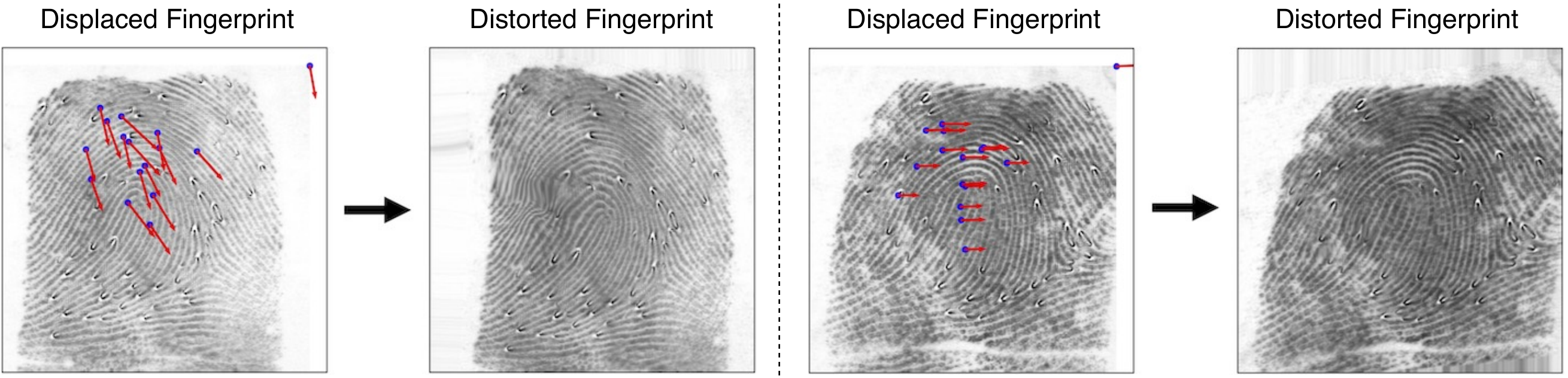}
  \captionsetup{font=footnotesize}
  \caption{Fingerprints in the left column are example displaced fingerprints from $\mathcal{G}_{disp}$. The distortion module $\mathcal{G}_{dist}$ predicts control points (marked in blue) and distorts the images based on their displacements (red arrows) using the non-linear distortion model from~\cite{nonlineardistortion}. The resultant distorted fingerprint images are shown in the right column.}
  \label{fig:dist}
\end{figure}

Figure \ref{fig:dist} shows two displaced fingerprints and their corresponding output from $\mathcal{G}_{dist}$.

\paragraph{Discriminator}

In order to guide the generative modules $\mathcal{G}_{disp}$ and $\mathcal{G}_{dist}$ to synthesize realistic fingerprint images, we introduce a fully convolutional network as a patch-based discriminator $\mathcal{D}$. The job of the discriminator is to distinguish between real fingerprint images $x$ and the generated adversarial fingerprint images $x_{adv}$. This is accomplished through the GAN loss:

\begin{gather}
    \mathcal{L}_{gan} = log\mathcal{D}(x) + log(1 - \mathcal{D}(x_{adv}))
\end{gather}

The proposed approach \NAME is trained in an end-to-end manner with respect to the following objective function:

\begin{dmath}
    \mathcal{L} = \mathcal{L}_{gan} + \lambda_{mmap\_sim}\mathcal{L}_{mmap\_sim} + \lambda_{mmap\_dis}\mathcal{L}_{mmap\_dis} + \lambda_{pixel}\mathcal{L}_{pixel}
\end{dmath}

where the hyper-parameters $\lambda_{mmap\_sim}$, $\lambda_{mmap\_dis}$, and $\lambda_{pixel}$ denote the relative importance of their respective losses. Once trained, \NAME can generate an adversarial fingerprint image for any input probe fingerprint and can be tested on any fingerprint matcher regardless of the feature extraction method (minutiae or deep-features).

\begin{figure*}[t]
  \includegraphics[width=\linewidth]{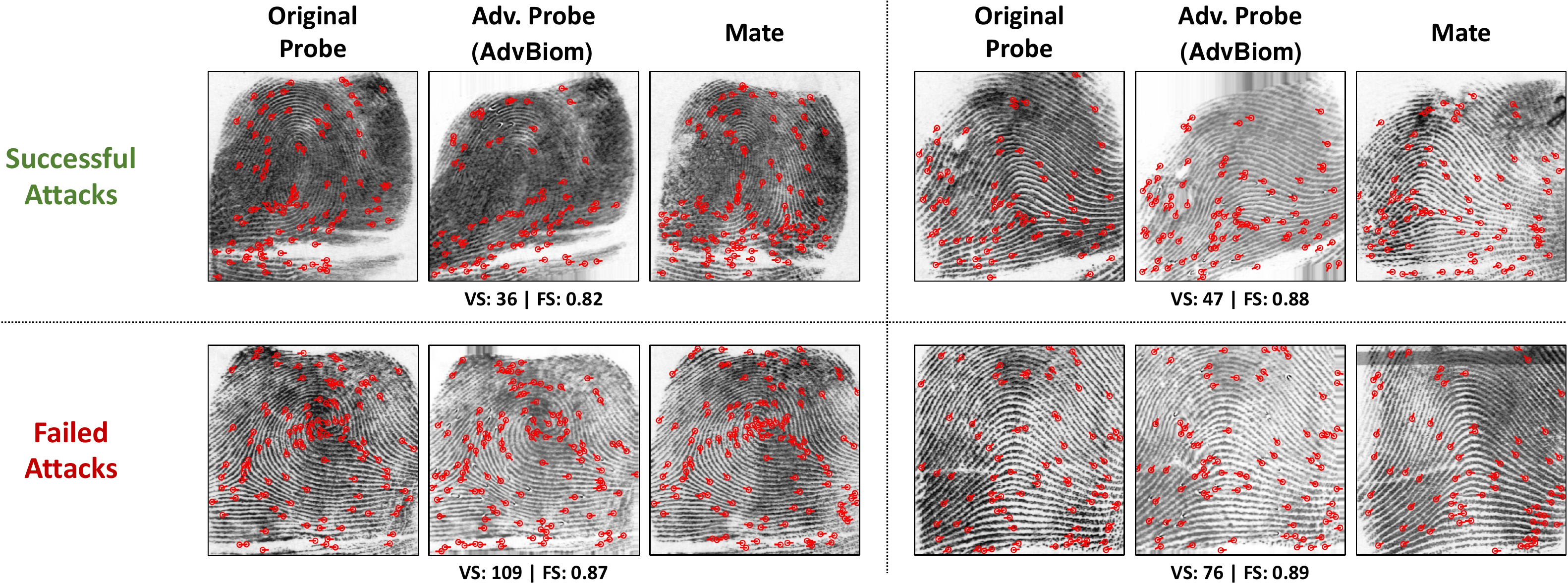}
  \captionsetup{font=footnotesize}
  \caption{Example successful and failed adversarial fingerprints attack using \NAME on NIST SD4~\cite{nist_sd4}. The VeriFinger matching scores (probe v. mate): VS, and fingerprintness~\cite{fingerprintness} scores: FS, of adversarial probes are shown below their respective triplet. Note that the VeriFinger matching threshold is 48 at 0.01\% FAR.}
  \label{fig:examples}
\end{figure*}

\noindent\textbf{\textit{Evaluation Metrics: }} The requirement of a good adversarial fingerprints generator is to evade fingerprint matchers while preserving fingerprint attributes and being model-agnostic. Thus, in order to quantify the performance of adversarial attacks generated by \NAME and other state-of-the-art baselines, we employ the following evaluation metrics:

\begin{itemize}
    \item \textit{True Accept Rate (TAR)}: The extent to which an adversarial attack can evade a fingerprint matcher is measured by the drop in TAR at an operational setting, say 0.01\% False Accept Rate (FAR).
    \item \textit{Fingerprintness}: Soweon and Jain~\cite{fingerprintness} proposed a domain-specific metric called fingerprintness to measure the degree of similarity of a given image to a fingerprint pattern. Fingerprintness ranges from [0,1] where higher the score, higher the probability of the pattern in the image corresponding to a fingerprint pattern.
    \item \textit{NFIQ 2.0}: Lastly, we use NFIQ 2.0~\cite{nfiq} quality scores to evaluate the fingerprint quality of adversarial fingerprint images. NFIQ scores range from [0,100] where a score of 100 depicts the highest fingerprint quality.
\end{itemize}

\begin{table}[!t]
\centering
\def\arraystretch{1.5}
\begin{tabular}{ccc|cccccc}
\hline
\multicolumn{3}{c|}{\multirow{2}{*}{\textbf{Accuracy}}} & \multicolumn{6}{c}{\textbf{Adversarial Attacks}} \\ \cline{4-9} 
\multicolumn{3}{c|}{} & \multicolumn{1}{c|}{\textbf{\begin{tabular}[c]{@{}c@{}}Original \\ Probes\end{tabular}}} & \multicolumn{1}{c|}{\textbf{FGSM}} & \multicolumn{1}{c|}{\textbf{I-FGSM}} & \multicolumn{1}{c|}{\textbf{\begin{tabular}[c]{@{}c@{}}Deep\\ Fool\end{tabular}}} & \multicolumn{1}{c|}{\textbf{PGD}} & \textbf{\begin{tabular}[c]{@{}c@{}}Adv\\ Biom\end{tabular}} \\ \hline
\multicolumn{1}{c|}{\multirow{9}{*}{\textbf{\begin{tabular}[c]{@{}c@{}}TAR\\ (\%)\\  at \\ 0.01\%\\ FAR\end{tabular}}}} & \multicolumn{1}{c|}{\multirow{3}{*}{\textbf{\begin{tabular}[c]{@{}c@{}}NIST\\ SD4\end{tabular}}}}           & \textbf{VeriFinger}  & \multicolumn{1}{c|}{99.05}                                                               & \multicolumn{1}{c|}{95.20}         & \multicolumn{1}{c|}{98.30}           & \multicolumn{1}{c|}{95.00}                                                        & \multicolumn{1}{c|}{97.60}        & \textbf{56.25}                                               \\ \cline{3-3}
\multicolumn{1}{c|}{}                                                                                               & \multicolumn{1}{c|}{}                                                                                       & \textbf{Innovatrics} & \multicolumn{1}{c|}{97.00}                                                               & \multicolumn{1}{c|}{93.00}         & \multicolumn{1}{c|}{95.50}           & \multicolumn{1}{c|}{92.65}                                                        & \multicolumn{1}{c|}{94.75}        & \textbf{41.35}                                               \\ \cline{3-3}
\multicolumn{1}{c|}{}                                                                                               & \multicolumn{1}{c|}{}                                                                                       & \textbf{DeepPrint}   & \multicolumn{1}{c|}{94.55}                                                               & \multicolumn{1}{c|}{36.20}         & \multicolumn{1}{c|}{64.15}           & \multicolumn{1}{c|}{\textbf{30.40}}                                               & \multicolumn{1}{c|}{68.75}        & 46.35                                                        \\ \cline{2-9} 
\multicolumn{1}{c|}{}                                                                                               & \multicolumn{1}{c|}{\multirow{3}{*}{\textbf{\begin{tabular}[c]{@{}c@{}}NIST\\ SD14\end{tabular}}}}          & \textbf{VeriFinger}  & \multicolumn{1}{c|}{99.42}                                                               & \multicolumn{1}{c|}{95.20}         & \multicolumn{1}{c|}{98.30}           & \multicolumn{1}{c|}{95.00}                                                        & \multicolumn{1}{c|}{97.60}        & \textbf{37.67}                                               \\ \cline{3-3}
\multicolumn{1}{c|}{}                                                                                               & \multicolumn{1}{c|}{}                                                                                       & \textbf{Innovatrics} & \multicolumn{1}{c|}{98.24}                                                               & \multicolumn{1}{c|}{90.84}         & \multicolumn{1}{c|}{95.68}           & \multicolumn{1}{c|}{91.32}                                                        & \multicolumn{1}{c|}{94.01}        & \textbf{25.69}                                               \\ \cline{3-3}
\multicolumn{1}{c|}{}                                                                                               & \multicolumn{1}{c|}{}                                                                                       & \textbf{DeepPrint}   & \multicolumn{1}{c|}{96.52}                                                               & \multicolumn{1}{c|}{48.70}         & \multicolumn{1}{c|}{84.48}           & \multicolumn{1}{c|}{\textbf{31.44}}                                               & \multicolumn{1}{c|}{64.28}        & 69.42                                                        \\ \cline{2-9} 
\multicolumn{1}{c|}{}                                                                                               & \multicolumn{1}{c|}{\multirow{3}{*}{\textbf{\begin{tabular}[c]{@{}c@{}}FVC\\ 2004\\ DB1 A\end{tabular}}}} & \textbf{VeriFinger}  & \multicolumn{1}{c|}{94.89}                                                               & \multicolumn{1}{c|}{91.60}         & \multicolumn{1}{c|}{91.53}           & \multicolumn{1}{c|}{86.92}                                                        & \multicolumn{1}{c|}{92.69}        & \textbf{22.31}                                               \\ \cline{3-3}
\multicolumn{1}{c|}{}                                                                                               & \multicolumn{1}{c|}{}                                                                                       & \textbf{Innovatrics} & \multicolumn{1}{c|}{94.15}                                                               & \multicolumn{1}{c|}{87.36}         & \multicolumn{1}{c|}{85.68}           & \multicolumn{1}{c|}{82.32}                                                        & \multicolumn{1}{c|}{88.75}        & \textbf{5.52}                                                \\ \cline{3-3}
\multicolumn{1}{c|}{}                                                                                               & \multicolumn{1}{c|}{}                                                                                       & \textbf{DeepPrint}   & \multicolumn{1}{c|}{75.36}                                                               & \multicolumn{1}{c|}{13.22}         & \multicolumn{1}{c|}{33.31}           & \multicolumn{1}{c|}{\textbf{6.87}}                                                & \multicolumn{1}{c|}{27.39}        & 20.62                                                        \\ \hline
\end{tabular}
\captionsetup{font=footnotesize}
\caption{True Accept Rate (TAR) @ 0.01\% FAR of \NAME along with state-of-the-art baselines attacks on three datasets - NIST SD4~\cite{nist_sd4}, NIST SD14~\cite{nist_sd14}, and FVC 2004 DB1 A~\cite{fvc2004}. 2 COTS fingerprint matchers - VeriFinger v11.0~\cite{verifinger} and Innovatrics v7.6.0.627~\cite{innovatrics}, and a deep network-based matcher DeepPrint~\cite{deepprint} were employed for the evaluation. It is observed that DeepPrint, a deep network-based matcher, is susceptible to all types of adversarial attacks while VeriFinger and Innovatrics are more robust.}
    \label{tab:tar_attacks}
\end{table}

Note that since non-linear distortions change the structure of the image, using the structural similarity index (SSIM) metric is inappropriate as it essentially measures the local change in structures of the image pairs.

\bigbreak
\noindent\textbf{\textit{Datasets: }} We train \NAME on an internal dataset of ~120,000 rolled fingerprint images. Furthermore, we evaluate the performance of the proposed fingerprint adversarial attack and other baselines on:

\begin{itemize}
    \item 2,000 fingerprint pairs from NIST SD4~\cite{nist_sd4}
    \item 27,000 fingerprint pairs from NIST SD14~\cite{nist_sd14}
    \item 558 fingerprints from DB1 A of FVC 2004~\cite{fvc2004}, consisting of 1,369 genuine pairs.
\end{itemize}

\bigbreak
\noindent\textbf{\textit{Experimental Settings: }} \NAME was trained using the Adam optimizer with $\beta_{1}$ as 0.5 and $\beta_{2}$ as 0.9. The hyper-parameters were empirically set to $\lambda_{mmap\_sim}=0.05$, $\lambda_{mmap\_dis}=500000$, and $\lambda_{pixel}=1000$ for convergence. Based on the conclusions drawn in~\cite{whitebox}, $d$, $c$, and $\lambda$ were set to 20, 16, and 2.0 respectively for optimal effectiveness against fingerprint matchers while ensuring fingerprint realism. \NAME was trained for 16,000 steps using Tensorflow r1.14.0 on an Intel Core i7-11700F @ 2.50GHz CPU with a RTX 3070 GPU. On the same machine, \NAME can synthesize an adversarial fingerprint within 0.35 seconds.\\

\noindent\textbf{\textit{Fingerprint Authentication Systems: }} Since \NAME is a black-box attack, we do not require any fingerprint authentication system while training the network. However, we evaluate \NAME and other baseline attacks on two COTS fingerprint matchers and one deep network-based matcher:

\begin{itemize}
    \item VeriFinger v11.0~\cite{verifinger}
    \item Innovatrics v7.6.0.627~\cite{innovatrics}
    \item DeepPrint~\cite{deepprint}
\end{itemize}

\paragraph{Comparison with Prevailing Fingerprint Adversarial Generators}

We show the performance of our method \NAME as compared to other state-of-the-art attacks in Table~\ref{tab:tar_attacks}. We observe that the TAR of two COTS and a deep network-based fingerprint matcher for the aforementioned three datasets. It is to note that all the baseline attacks~\cite{fgsm, ifgsm, pgd, deepfool} are white-box attacks and were trained using DeepPrint~\cite{deepprint}. It is evident from Table~\ref{tab:tar_attacks} that \NAME is the most successful attack on COTS matchers VeriFinger and Innovatrics, and is also able to effectively evade a deep network-based fingerprint matcher, namely DeepPrint. It can also be observed that while COTS fingerprint matchers are robust to most adversarial attacks, DeepPrint is very susceptible to the same attacks since it heavily relies on the texture of the fingerprint which is majorly affected by adversarial attacks.

\begin{figure}
\centering
\begin{minipage}{.48\textwidth}
  \centering
  \includegraphics[width=\linewidth]{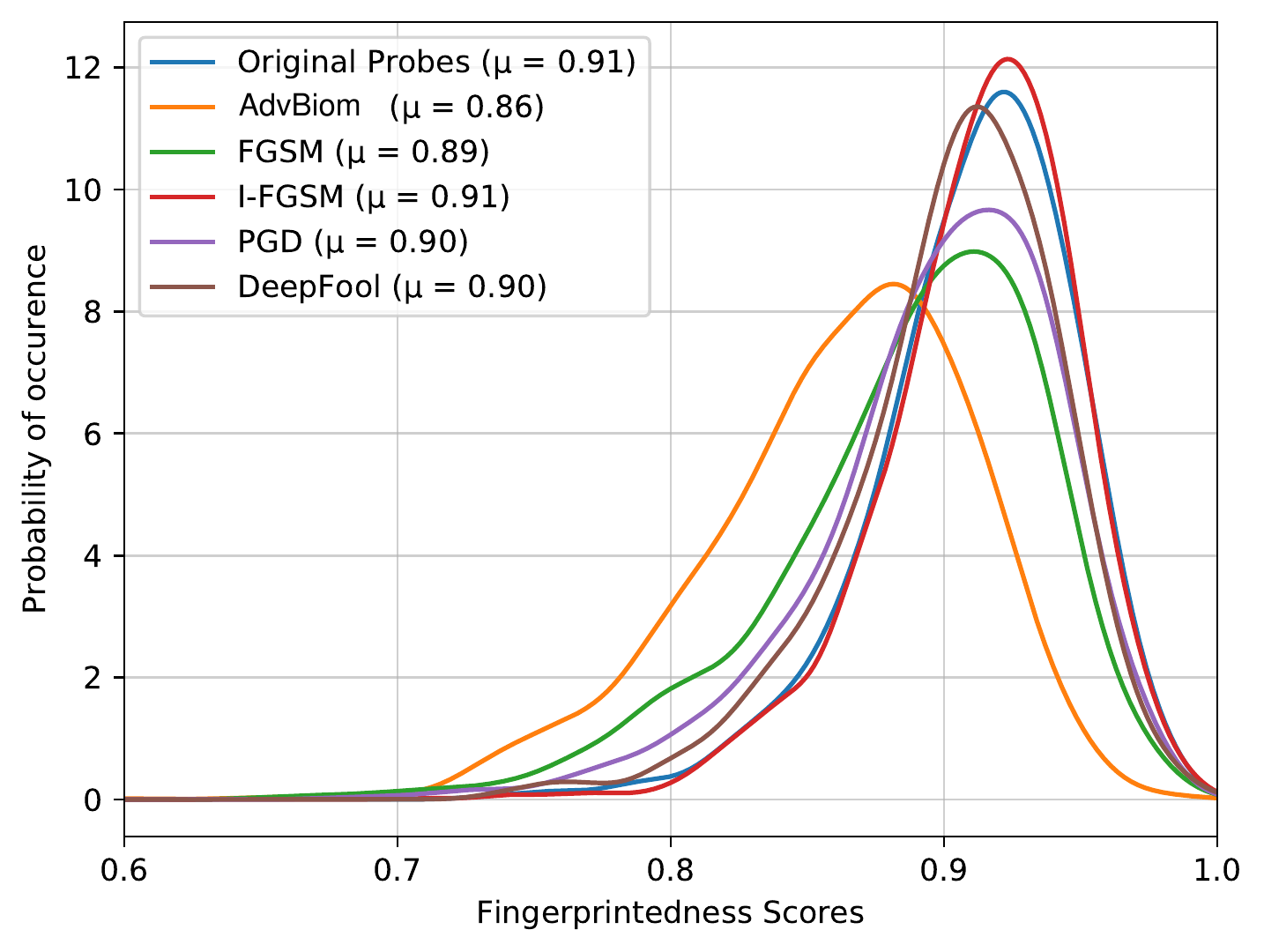}
  \captionsetup{font=footnotesize}
  \captionof{figure}{Fingerprintness~\cite{fingerprintness} distribution of 2,000 probes from NIST SD4 with respect to \NAME and other state-of-the-art baselines attacks.}
  \label{fig:fingerprintness}
\end{minipage}%
\hfill
\begin{minipage}{.48\textwidth}
  \centering
  \includegraphics[width=\linewidth]{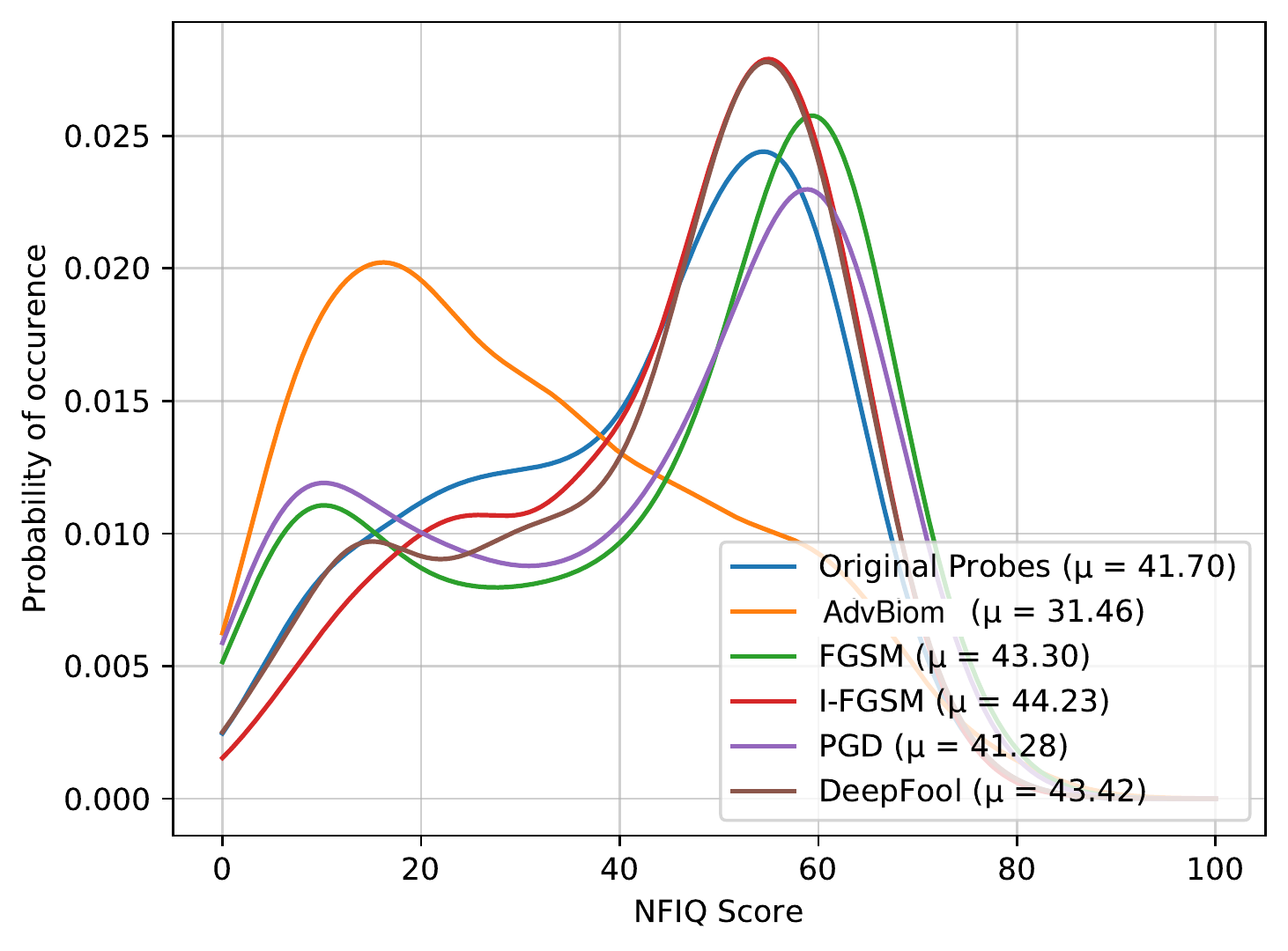}
  \captionsetup{font=footnotesize}
  \captionof{figure}{NFIQ 2.0~\cite{nfiq} quality scores distribution of 2,000 probes from NIST SD4~\cite{nist_sd4} with respect to \NAME and other baselines attacks.}
  \label{fig:nfiq}
\end{minipage}
\end{figure}

A successful adversarial attack should not only evade fingerprint matchers but should also preserve fingerprint attributes. In order to observe the effect of adversarial attacks on fingerprint pattern in images, we plot the fingerprintness~\cite{fingerprintness} distribution of 2,000 probes from NIST SD4~\cite{nist_sd4} for \NAME as well as for other baseline attacks. Since all the state-of-the-art baselines essentially add noise to each pixel in the image, they do not change the structure of the fingerprint and thus do not affect fingerprintness scores. \NAME, on the other hand, displaces minutiae points and non-linearly distorts the image, and still maintains a high mean fingerprintness score of $\mu = 0.86$.

Furthermore, we also compute the NFIQ 2.0~\cite{nfiq} quality scores distribution (figure~\ref{fig:nfiq}) of the original and adversarial probes from NIST SD4~\cite{nist_sd4}. As shown in figure~\ref{fig:example_attack}, baseline attacks tend to minutely perturb image pixels to generate adversarial fingerprints and as a result do not have much of an effect on the quality scores. \NAME, on the other hand, \textbf{provides an optimal solution} by successfully attacking fingerprint matchers while maintaining high fingerprintness and NFIQ scores.

\paragraph{Genuine and Imposter Scores Distribution}

\begin{figure*}[!h]
\centering
\resizebox{\textwidth}{!}{\begin{subfigure}{.33\textwidth}
  \centering
  \includegraphics[width=\linewidth]{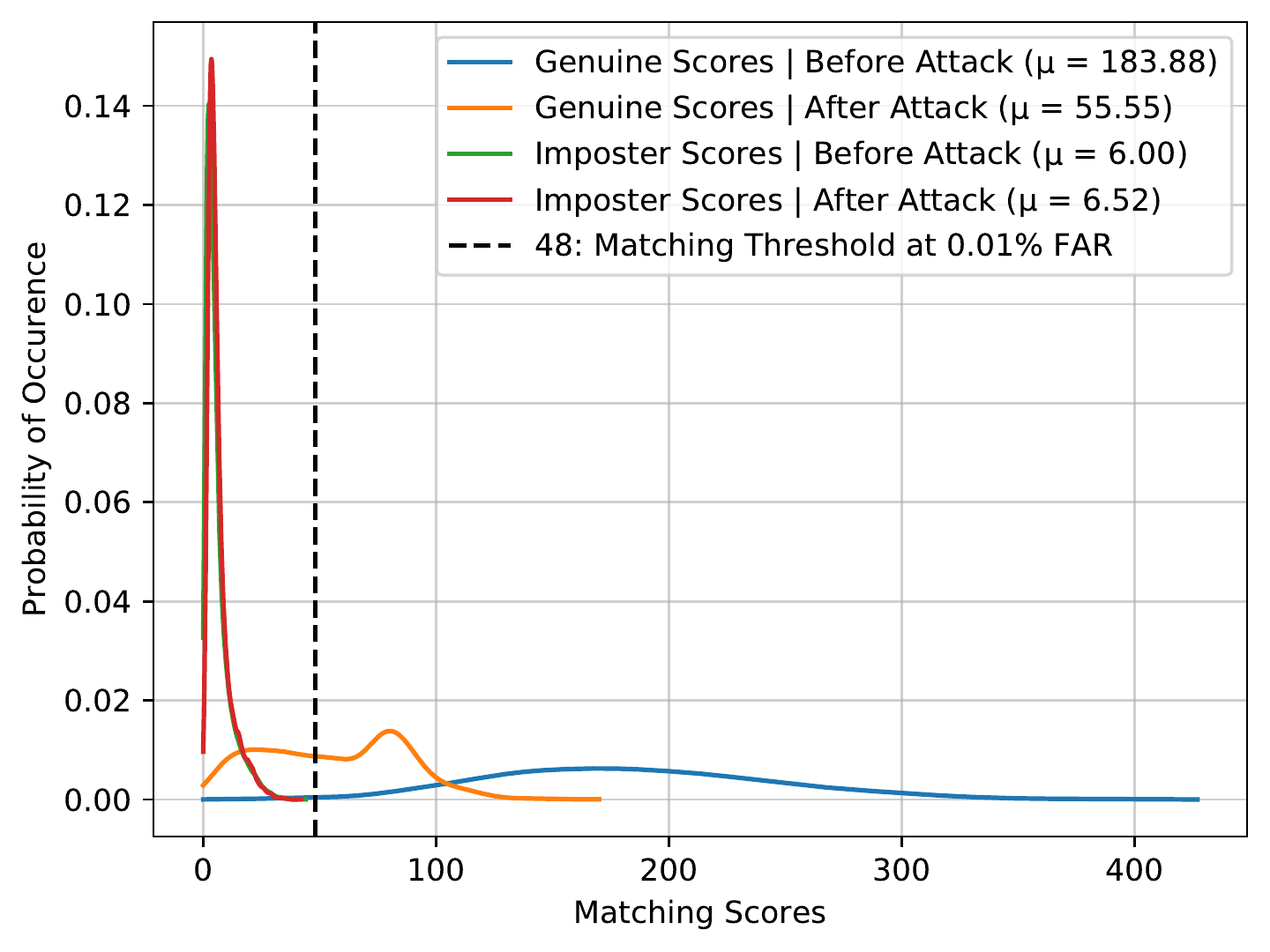}
  \captionsetup{font=footnotesize}
  \caption{Using VeriFinger SDK~\cite{verifinger}}
  \label{fig:gen_imp_nist4_ver}
\end{subfigure}%
\begin{subfigure}{.33\textwidth}
  \centering
  \includegraphics[width=\linewidth]{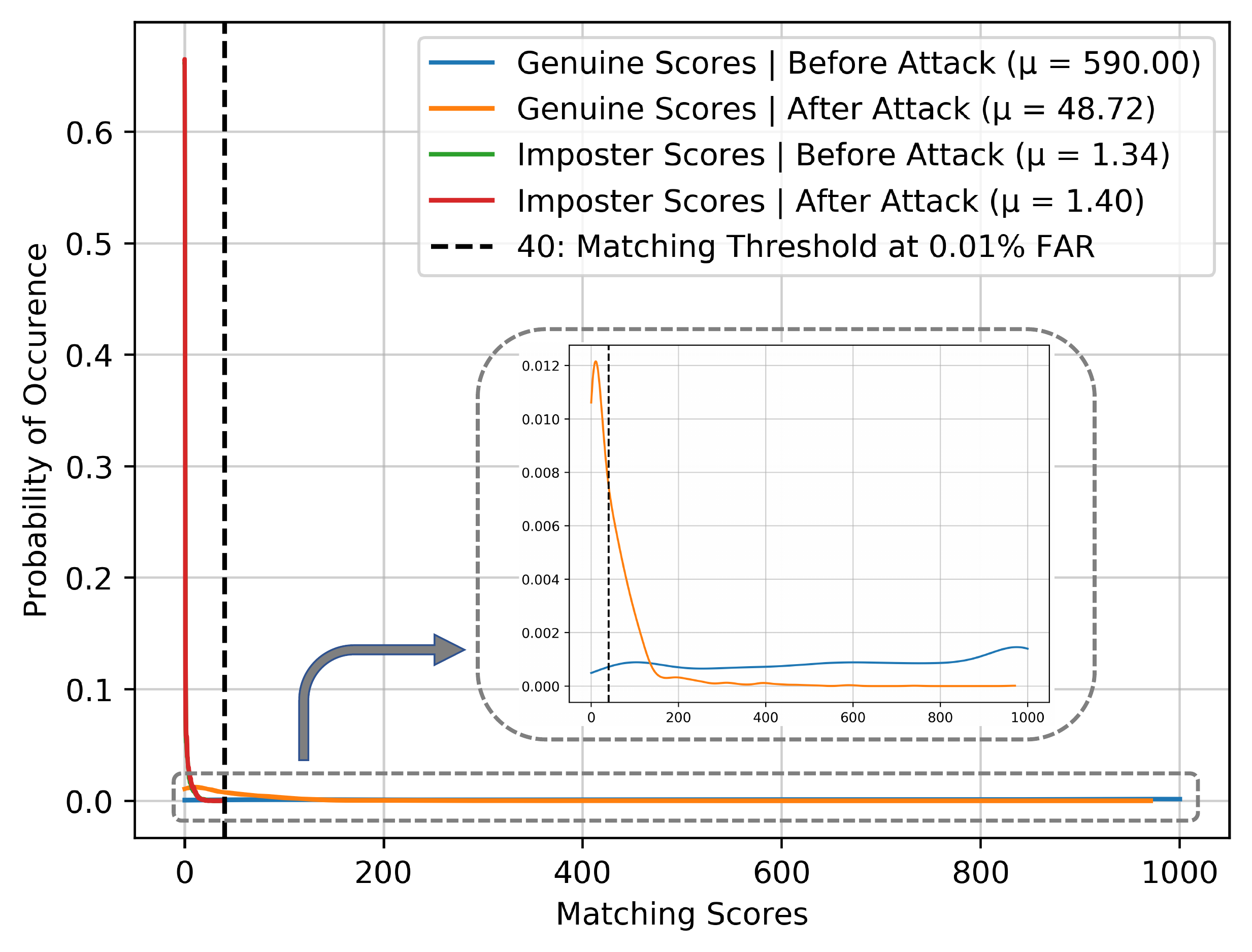}
  \captionsetup{font=footnotesize}
  \caption{Using Innovatrics SDK~\cite{innovatrics}}
  \label{fig:gen_imp_nist4_inn}
\end{subfigure}

\begin{subfigure}{.33\textwidth}
  \centering
  \includegraphics[width=\linewidth]{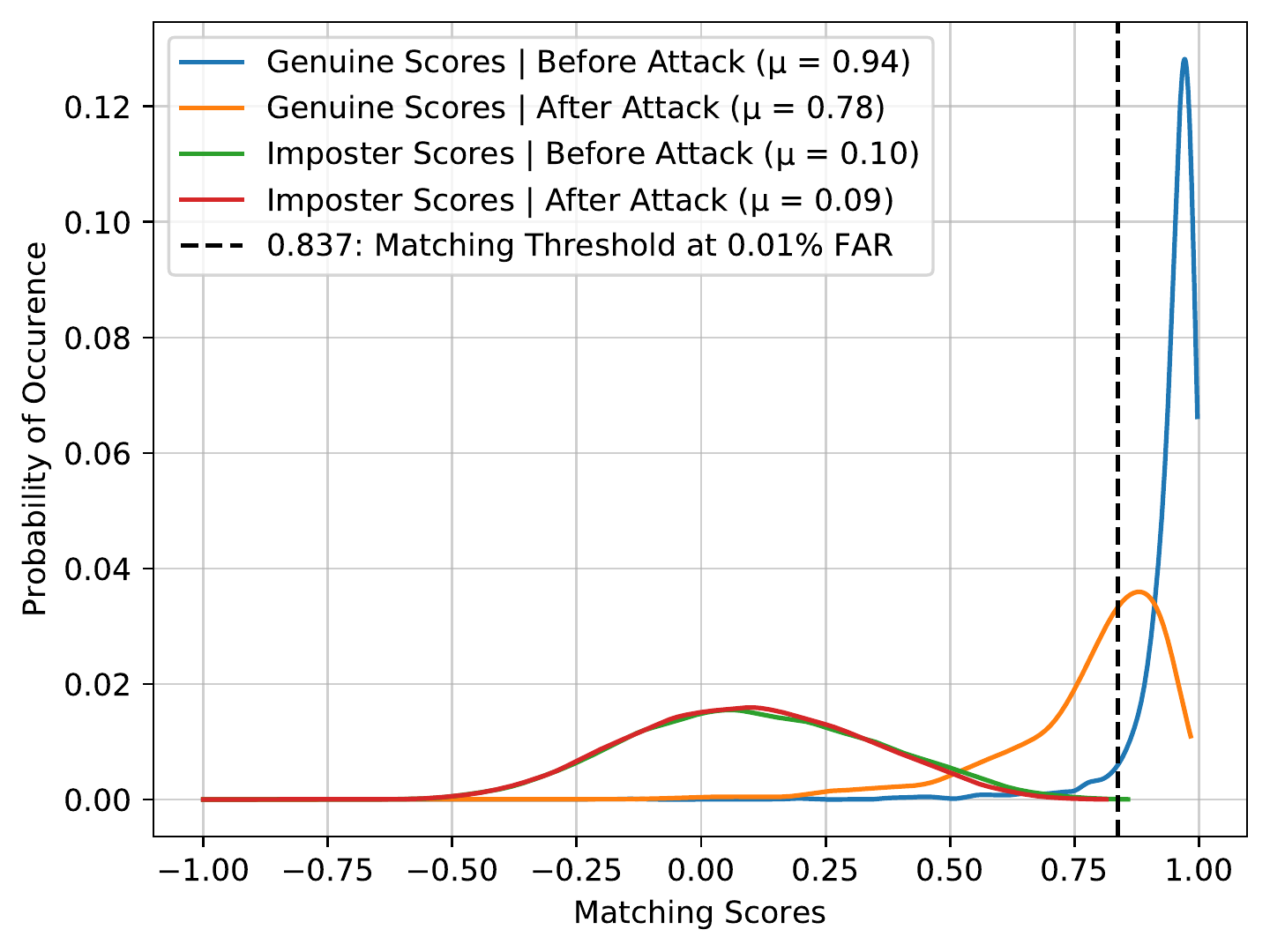}
  \captionsetup{font=footnotesize}
  \caption{using DeepPrint~\cite{deepprint}}
  \label{fig:gen_imp_nist4_dp}
\end{subfigure}}
\captionsetup{font=footnotesize}
\caption{Genuine and imposter scores distribution of NIST SD4~\cite{nist_sd4} before and after the adversarial attack \NAME using three state-of-the-art fingerprint matchers - VeriFinger v11.0~\cite{verifinger}, Innovatrics v7.6.0.627~\cite{innovatrics}, and DeepPrint~\cite{deepprint}. Here, $\mu$ refers to the mean of the scores distribution. In all the three cases, the genuine scores shift towards the left while the imposter scores do not get affected by the attack.}
\label{fig:gen_imp_nist4}
\end{figure*}

To determine the effect of adversarial fingerprint on both genuine and imposter pairs, we plot the genuine and imposter scores distribution of NIST SD4~\cite{nist_sd4} in figure~\ref{fig:gen_imp_nist4} before and after applying \NAME. We computed a total of 2,000 genuine and 20,000 imposter scores for the evaluation. It can be observed that the genuine scores drastically decrease and shift to the left of the axis as their mean drops from 183.87 to 55.55 after the attack. However, the imposter scores remain unaffected with the mean imposter score changing by only 0.53.

\paragraph{Is \NAME Biased Towards Certain Fingerprint Types?}

The generated adversarial fingerprint from \NAME is conditioned on the input probe fingerprint. Thus, it is essential to check if there is a relation between the amount of perturbation applied and the fingerprint type. The confusion matrix for the five fingerprint types (left loop, right loop, whorl, arch, tented arch) before and after applying \NAME on the 2,000 probes of NIST SD4~\cite{nist_sd4} is shown in table~\ref{tab:conf_tab_type}. Note that we use NIST SD4 for this evaluation since it has a uniform number of fingerprint images per each type (400 fingerprints per type). It is evident from the table that all five fingerprint types are almost equally susceptible to the attack, and thus the attack crafted \NAME is not biased towards a particular fingerprint type.

\begin{table}[t]
\small
    \centering
    \def\arraystretch{1.5}
    \begin{tabular}{p{1.5cm}|p{1cm}|p{1.5cm}|p{1.2cm}||p{1.5cm}|p{1cm}|p{1.5cm}|p{1.2cm}}
    \Xhline{2\arrayrulewidth}
    \multicolumn{4}{c||}{\textbf{Before Attack}} & \multicolumn{4}{c}{\textbf{After Attack}}\\
    \Xhline{2\arrayrulewidth}
    \makecell{\textbf{TAR} \\L: 99.75\%\\R: 99.25\%\\W: 99.50\%\\T: 99.25\%\\A: 97.50\%} & 
    \makecell{\textbf{FAR} \\L: 0\%\\R: 0\%\\W: 0\%\\T: 0\%\\A: 0\%} &
    \makecell{\textbf{FRR} \\L: 0.25\%\\R: 0.75\%\\W: 0.50\%\\T: 0.75\%\\A: 2.50\%} &
    \makecell{\textbf{TRR} \\L: 100\%\\R: 100\%\\W: 100\%\\T: 100\%\\A: 100\%} &
    \makecell{\textbf{TAR} \\L: 59.00\%\\R: 56.50\%\\W: 58.75\%\\T: 56.00\%\\A: 57.00\%} &
    \makecell{\textbf{FAR} \\L: 0\%\\R: 0\%\\W: 0\%\\T: 0\%\\A: 0\%} &
    \makecell{\textbf{FRR} \\L: 41.00\%\\R: 43.50\%\\W: 41.25\%\\T: 44.00\%\\A: 43.00\%} &
    \makecell{\textbf{TRR} \\L: 100\%\\R: 100\%\\W: 100\%\\T: 100\%\\A: 100\%}\\
    \Xhline{2\arrayrulewidth}
    \end{tabular}
    \captionsetup{font=footnotesize}
    \caption{Confusion matrix for five fingerprint types (left loop: L, right loop: R, whorl: W, tented arch: T, arch: A) from NIST SD4~\cite{nist_sd4} before and after the adversarial attack using \NAMENOSPACE. Here, TAR = True Accept Rate, FAR = False Accept Rate, FRR = False Reject Rate, and TRR = True Reject Rate. Note that the matching threshold was 48 at 0.01\% FAR using the COTS fingerprint matcher VeriFinger. \NAME is not biased towards any fingerprint type.}
    \label{tab:conf_tab_type}
\end{table}{}

\section{Conclusions}
We show that a new method of adversarial synthesis, namely \NAMENOSPACE, that automatically generates adversarial face images with imperceptible perturbations evading state-of-the-art biometric matchers. With the help of a GAN, and the proposed perturbation and identity losses, \NAME learns the set of pixel locations required by face matchers for identification and only perturbs those salient facial regions (such as eyebrows and nose). Once trained, \NAME generates high quality and perceptually realistic adversarial examples that are benign to the human eye but can evade state-of-the-art black-box face matchers, while outperforming other state-of-the-art adversarial face methods. Beyond faces, we show for the first time that such a method with the proposed Minutiae Displacement and Distortion Modules can also evade state-of-the-art automated fingerprint recognition systems. 

\bibliographystyle{ieeetr}
\bibliography{bibliography}

\end{document}